\documentclass[letterpaper]{article}
\usepackage[preprint]{aaai2027}
\usepackage[hyphens]{url}
\usepackage{graphicx}
\usepackage{natbib}
\usepackage{caption}
\usepackage{booktabs}
\usepackage{amsmath}
\usepackage{amssymb}
\usepackage{multirow}
\usepackage{pgfplots}

\usepackage[most]{tcolorbox}
\tcbset{
  promptbox/.style={
    enhanced,
    breakable,
    colback=white,
    colframe=black!70,
    boxrule=0.5pt,
    arc=2pt,
    left=5pt,
    right=5pt,
    top=5pt,
    bottom=5pt,
    before skip=4pt,
    after skip=6pt,
    fontupper=\small
  }
}

\usepgfplotslibrary{groupplots}
\usetikzlibrary{patterns}
\pgfplotsset{compat=1.18}

\title{CraftAlign: Feature-Grounded Evaluation and Revision Guidance for AI Stories}
\author{Yang Yang\textsuperscript{1}, Boyun Xu\textsuperscript{1}, Shaofeng Liang\textsuperscript{1}, Yun Han\textsuperscript{1}, Zining Zhong\textsuperscript{1}, Songning Lai\textsuperscript{1}, Kaishen Yuan\textsuperscript{1}, Yutao Yue\textsuperscript{1,2}\thanks{Corresponding author: yueyutao@hkust-gz.edu.cn}}
\affiliations{
\textsuperscript{1}The Hong Kong University of Science and Technology (Guangzhou), Guangzhou 511400, China\\
\textsuperscript{2}Institute of Deep Perception Technology, JITRI, Wuxi 214000, China
}

\begin{document}

\maketitle

\begin{abstract}
Large language models can now generate fluent and complete stories, yet many outputs still feel formulaic and unnatural because of clichés, over-explanation, linear causal progression, and stereotyped endings—an immediately recognizable “AI flavor.” Existing detection and evaluation methods often stop at source labels or holistic scores, while revision methods typically target predefined issues through localized edits, limiting their ability to support multiple plausible revision strategies or guide story-wide changes in information release, causal organization, and ending treatment. We introduce CraftAlign, a framework that aligns AI stories with the craft of human storytelling by both assessing Human/AI writing patterns and providing revision guidance. CraftAlign comprises two learned modules and an inference-time guidance pipeline. A feature estimator built on Qwen3.5-9B predicts 304 explicit writing features spanning style and narrative. A class-conditional energy model scores the resulting feature configuration against Human and AI writing patterns, conditioning on the original writing prompt when available. At inference time, CraftAlign applies schema-valid structured perturbations, selects changes that move the feature configuration toward the Human writing pattern, and converts them into natural-language guidance for a separate editor to rewrite the full story. Experiments show that CraftAlign accurately distinguishes Human and AI writing patterns and that its guidance outperforms revision baselines across editors and in a human study.

\end{abstract}

\begin{figure}[!t]
  \centering
  \includegraphics[width=\columnwidth]{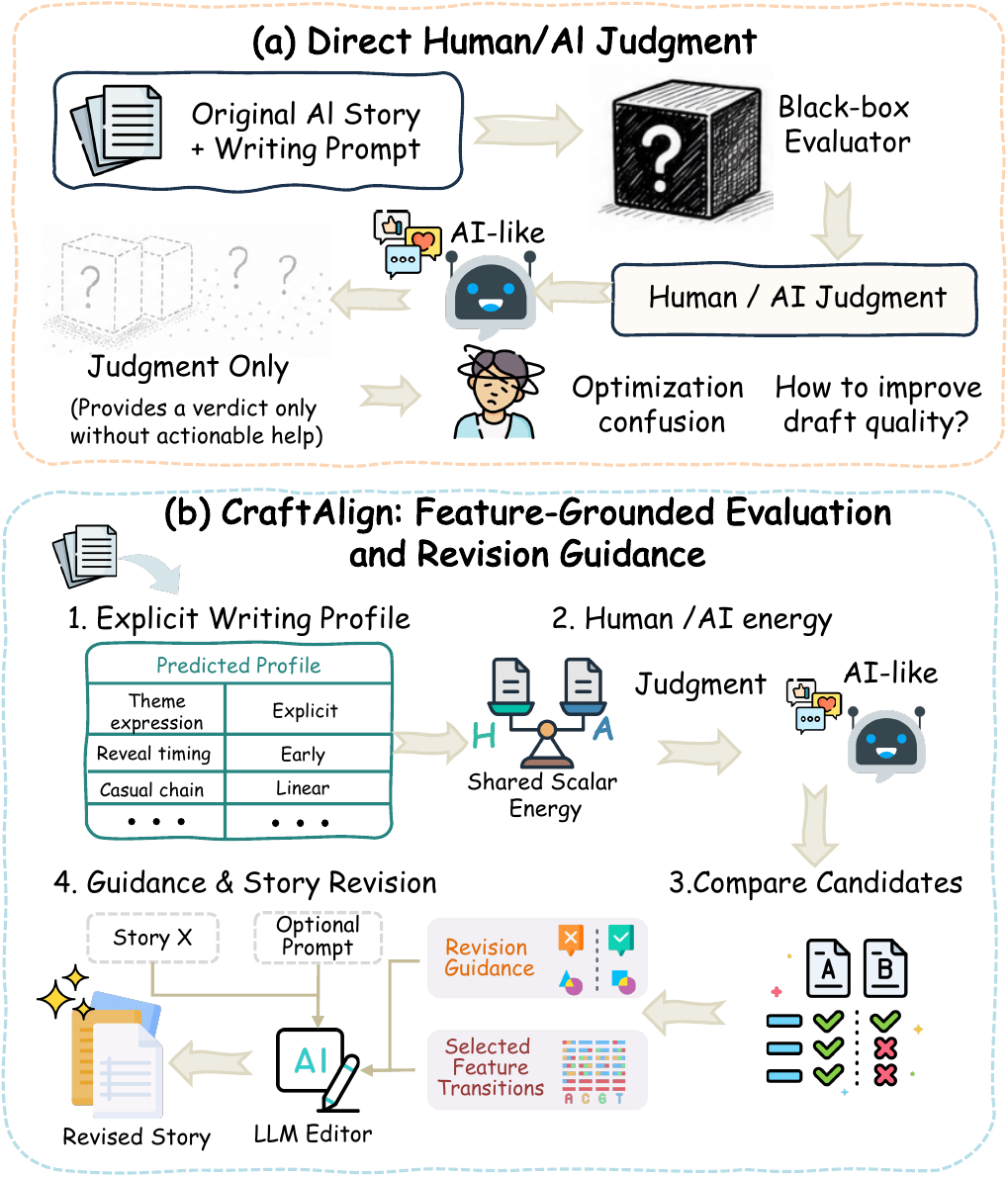}
  \caption{Comparison between direct Human/AI judgment and CraftAlign's
feature-grounded evaluation and revision.}
  \label{fig:intro}
\end{figure}

\section{Introduction}
Recent advances have made it increasingly easy for large language
models (LLMs) to generate stories that are fluent, coherent, and
complete. Yet producing a readable story is not the same as telling one
in the way a human author would. AI-generated fiction continues to
exhibit recurring differences in both style and narrative: clich\'es,
over-ornamentation, and redundant explanation affect local expression,
while explicit themes, overly linear causality, predictable information
release, and formulaic closure shape how events and information are
organized across the full story
\citep{chakrabarty2025salvaged,russell2026storyscope}. LLMs have
therefore narrowed the gap in whether a complete story can be produced,
but not in how that story is told.

Prior work has mainly approached these differences through source
detection or text revision. Detection methods can identify
machine-generated text from recurring stylistic regularities, but
usually provide a source judgment rather than actionable guidance for a
particular story
\citep{hans2024binoculars,sun2025idiosyncrasies,shaib2026slop}.
Revision methods can improve passages exhibiting predefined writing
artifacts, but less often address story-wide properties such as causal
organization, information release, and ending treatment
\citep{chakrabarty2025salvaged}. StoryScope provides a useful bridge
between these directions by representing each story through 304
explicit writing features spanning style and narrative. These features
support 96.0\% Macro-F1 in Human/AI source classification while
localizing the observed differences to interpretable dimensions
\citep{russell2026storyscope}. However,
StoryScope is primarily designed for analysis and diagnosis: it does
not determine which features of a particular story should be revised or
in which direction.

Turning feature-level diagnosis into revision guidance raises two
challenges. First, whether a writing pattern is appropriate may depend
on the original prompt. Explicit themes, linear progression, and
complete closure may suit a children's fable but feel formulaic in
psychological suspense; an evaluator that observes only writing features cannot directly
account for these contextual differences. Second, open-ended revision rarely
has a unique target. For example, suspense may be improved either by
delaying key revelations or by reducing explicit explanations of
motivation and causality. Both are plausible revisions, but they
correspond to different feature configurations. A useful
evaluation-and-guidance model must therefore condition on the writing
prompt when available, compare alternative feature changes, and select
promising directions without treating any single revised configuration
as the uniquely correct target.

To address these challenges, we introduce \textbf{CraftAlign}, a
feature-grounded framework that uses explicit writing features not only
to evaluate Human/AI writing patterns but also to guide full-story
revision. As illustrated in Figure~\ref{fig:intro}, CraftAlign treats
this feature space as a structured intervention space rather than
stopping at Human/AI judgment. A local Qwen3.5-9B estimator with
type-specific prediction heads predicts 304 heterogeneous writing
features spanning style and narrative. A prompt-optional,
class-conditional energy function then evaluates how well the predicted
feature configuration, together with the writing prompt when available,
matches Human and AI writing patterns \citep{lecun2006energy}. The
resulting Human--AI energy difference supports both writing-pattern
evaluation and comparison among alternative feature changes. At
inference time, CraftAlign fixes Human as the target, applies low-budget
structured perturbations to the predicted feature configuration, and
selects changes that move the story toward the Human writing pattern.
It then renders the selected changes as natural-language guidance for a
general-purpose LLM editor to rewrite the full story. Experiments show
that CraftAlign reliably evaluates Human/AI writing patterns with or
without the original prompt, while its targeted guidance outperforms
generic rewriting and count-matched random guidance across editors and
in human assessment. These findings also motivate a valuable next
direction: developing feature-grounded training signals that guide
story-generation models toward more human-like narrative patterns.

Our main contributions can be summarized as follows:
\begin{itemize}
    \item We develop \textbf{CraftAlign}, a deployable,
    feature-grounded framework for Human/AI writing-pattern evaluation.
    It combines a type-aware estimator of 304 explicit writing features
    spanning style and narrative with a prompt-optional,
    class-conditional energy model that scores compatibility with
    Human and AI writing patterns.

    \item We introduce a Human-targeted, energy-guided feature search
    strategy that generates schema-valid and text-actionable candidates
    and selects which writing features to target and in what direction.
    The selected feature transitions are rendered through a curated
    guidance dictionary into targeted revision guidance for full-story
    rewriting by a general-purpose LLM editor.

    \item We provide a systematic evaluation of the
    evaluation-to-revision pipeline, combining model-space analysis with
    human assessment and showing that feature-grounded evaluation
    supports effective full-story revision.
\end{itemize}

\section{Related Work}

\subsection{Evaluating and Revising AI-Generated Stories}

Prior work on AI-generated stories spans source identification,
creative-writing evaluation, and story generation or revision.
Source-identification methods use model scores or recurring writing
patterns to distinguish human- and machine-generated text, but
typically stop at a source label rather than explaining how a
particular story should be revised
\citep{hans2024binoculars,russell2025people}.
Creative-writing evaluators assign holistic scores or pairwise
preferences, yet do not decompose these judgments into searchable
changes in writing features
\citep{fein2026litbench,cao2026creval}.
Recent narrative benchmarking further shows that narrative events,
style, perspective, and revelation remain underrepresented in existing
evaluations, especially for subjective aspects without a single
correct answer \citep{hamilton2026narrabench}.
Long-form generation systems such as Re3 use planning, reranking, and
revision to improve plot coherence and premise relevance, while
artifact-focused revision methods locate and rewrite
expert-identified writing problems
\citep{yang2022re3,chakrabarty2025salvaged}.
However, their revision targets are not selected by comparing
alternative directions in an explicit writing-feature space.
StoryScope provides such a space by representing Human and AI stories
through explicit features spanning style and narrative, but uses these
features primarily for analysis and source classification
\citep{russell2026storyscope}.
CraftAlign instead generates and evaluates candidate transitions in
this feature space, selects directions that better match the target
writing pattern, and renders the selected transitions as story-level
revision guidance.

\subsection{Concept-Based Modeling and Intervention}

Concept bottleneck models (CBMs) predict human-interpretable concepts
before making downstream decisions, thereby supporting model
inspection and concept-level intervention \citep{koh2020concept}.
Subsequent work has enriched concept representations and studied more
effective intervention procedures and intervention-aware training
strategies
\citep{zarlenga2022cem,shin2023intervention,zarlenga2023intcem}.
Related reward-modeling approaches similarly decompose an overall
preference signal into explicit intermediate objectives or concepts
\citep{wang2024armorm,laguna2025cbrm}.
CraftAlign also relies on an explicit intermediate representation, but
assigns it a different role: its writing features define the structured
space of possible story revisions. Many concept-intervention methods assume externally provided corrected
concept values, whereas open-ended story revision provides neither a
unique ground-truth feature configuration nor a single correct
intervention. CraftAlign therefore treats the feature layer as
a target-label-driven search space, evaluates multiple schema-valid and
text-actionable transitions, and selects directions that better match
the target writing pattern.

\subsection{Energy-Based Modeling and Feature Search}
Energy-based models (EBMs) assign scalar energies to variable
configurations, with lower energy generally indicating greater
compatibility with the data or task constraints
\citep{lecun2006energy}. Energy-based objectives have also been used
to steer text generation directly during decoding
\citep{qin2022cold}. CraftAlign instead applies a class-conditional
energy function to an explicit writing-feature space, using the same
scoring interface for Human/AI writing-pattern evaluation and, after
fixing Human as the target, for comparing candidate feature
transitions. To generate schema-valid candidates, CraftAlign adapts
the type-aware local perturbations introduced for discrete and mixed
variables by \citet{schroder2024heat}. The resulting search is also
related to counterfactual explanation and algorithmic recourse, which
identify actionable changes leading to a target prediction, including
methods that generate multiple diverse alternatives
\citep{wachter2017counterfactual,ustun2019recourse,
mothilal2020dice,karimi2022recourse}. In the story domain, however,
a story-feature transition cannot be applied directly to the text:
CraftAlign must render the selected transition as natural-language
guidance and rely on a generative editor to realize it in the complete
story. CraftAlign therefore connects writing-pattern evaluation,
feature-space search, and story-level revision guidance through a
shared energy signal, while its feature-level auditability comes from
explicit writing features, single-feature transitions, and their
counterfactual energy changes rather than the energy formulation alone.

\begin{figure*}[t]
  \centering
  \includegraphics[width=\textwidth]{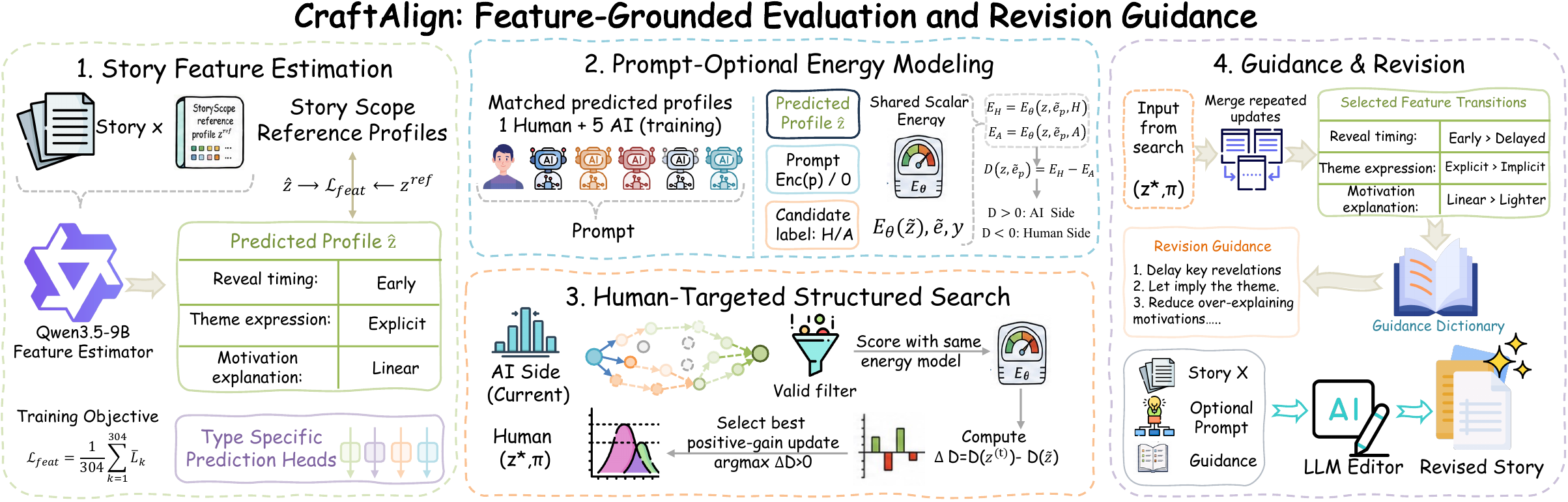}
  \caption{CraftAlign overview. StoryScope reference profiles supervise only the local feature estimator. Downstream energy modeling and search operate on out-of-sample predicted profiles. The same scalar energy function is queried with Human and AI labels, and structured search reuses this function to select renderable feature transitions.}
  \label{fig:method}
\end{figure*}

\section{Method}

\label{sec:method}

We formulate feature-grounded Human/AI writing-pattern evaluation
and story revision guidance in the explicit writing-feature space
introduced by StoryScope, which comprises 304 features spanning style
and narrative \citep{russell2026storyscope}. Rather than using these
features only as diagnostic inputs, CraftAlign treats them as a
structured intervention space in which alternative revision directions
can be evaluated and selected. Given a story \(x\) and an optional
writing prompt \(p\), CraftAlign maps the story to a predicted feature
configuration \(\hat{z}\), evaluates the resulting writing pattern under
candidate Human and AI labels, and, with Human fixed as the target,
derives feature-level guidance \(h\) for an editor that produces a
revised story \(x'\). Figure~\ref{fig:method} summarizes the full
pipeline and makes its deployable data flow explicit.

\subsection{Story Feature Estimation}
\label{sec:feature-estimation}

To enable reproducible and scalable application of the StoryScope
feature schema to new stories, we fine-tune Qwen3.5-9B together with
304 feature-specific prediction heads using the released
story--reference-profile pairs
\((x,z^{\mathrm{ref}})\) as supervision. Given a story \(x\), we use
the final-layer hidden state of the last sequence token as a shared
story representation and feed it to a separate head for each feature,
yielding \(g_\phi:x\mapsto\hat{z}\). The 304 features have
heterogeneous categorical, binary, ordinal, scale, and multi-select
output structures. Categorical features use softmax classification,
binary features use sigmoid classification, and multi-select features
use independent sigmoid outputs over their legal options. Ordinal
features and non-metric ordered scales are modeled with CORN
\citep{shi2023corn}, whereas metric scales use Huber regression.

Different features decompose into different numbers of thresholds or
legal options, so directly summing all output losses would overweight
features with larger output spaces. We therefore first average the loss
within each feature and then average equally across the 304 features:
\[
\mathcal{L}_{\mathrm{feat}}
=
\frac{1}{304}
\sum_{k=1}^{304}
\bar{\mathcal{L}}_k,
\]
where \(\bar{\mathcal{L}}_k\) is the loss for feature \(k\), averaged
over its valid output units. The resulting outputs are hard-decoded
into schema-valid feature values. After training, \(g_\phi\) is kept
fixed and used to generate predicted profiles for all downstream
stories and, during revision evaluation, their rewritten versions.
Thus, energy-model training, evaluation, and search all operate in the
same predicted-profile space available at inference, rather than using
StoryScope reference annotations. Prediction heads, decoding rules,
schema validation, and training settings are detailed in Appendix~C;
data isolation and out-of-sample profile generation are described in
Appendices~A and D.1.

\subsection{Prompt-Optional Group-Aware Energy Modeling}
\label{sec:energy-modeling}

For each writing prompt \(p_i\), StoryScope provides a matched group
containing one human-authored story \(x_i^H\) and five AI-generated
stories \(\{x_{ij}^A\}_{j=1}^{5}\). The frozen feature estimator maps
these stories to predicted profiles
\(\hat{z}_i^H=g_\phi(x_i^H)\) and
\(\hat{z}_{ij}^A=g_\phi(x_{ij}^A)\), respectively. Although every
training group has an associated prompt, we train a single model to
support both prompt-conditioned and prompt-free inference. For each
group, we therefore construct two condition representations:
\(\operatorname{Enc}(p_i)\) for the prompt-conditioned mode and an
all-zero vector \(\mathbf{0}\) of the same dimension for the
prompt-masked mode. The zero vector indicates that prompt information
is deliberately withheld and carries no prompt semantics.

The class-conditional energy model takes a predicted profile
\(\hat{z}\), a condition representation \(\tilde{e}_p\), and a
candidate label \(y\in\{H,A\}\), and assigns the resulting
configuration a scalar energy:
\[
\begin{aligned}
E_\theta(\hat{z},\tilde{e}_p,y)
&=
\operatorname{MLP}_\theta
\left(
\left[
\operatorname{EncFeature}(\hat{z});
\tilde{e}_p;
\operatorname{Emb}(y)
\right]
\right)
\in\mathbb{R},
\\
D(\hat{z},\tilde{e}_p)
&=
E_\theta(\hat{z},\tilde{e}_p,H)
-
E_\theta(\hat{z},\tilde{e}_p,A).
\end{aligned}
\]
Lower energy indicates a better match to the queried label.
Accordingly, \(D<0\) favors Human, whereas \(D>0\) favors AI. This
shared compatibility function is central to CraftAlign: the same scalar
interface supports both label comparison and counterfactual feature
search without requiring a separate scorer for revision directions.

Training uses class-balanced pointwise supervision as the primary
objective and within-prompt listwise supervision as an auxiliary
objective. The pointwise margin loss learns standalone Human/AI
judgments while assigning half of each group's total weight to the
human story and the other half collectively to the five AI stories,
preventing the \(1{:}5\) group composition from dominating training.
Following listwise learning-to-rank formulations
\citep{cao2007listnet}, the listwise softmax loss ranks the six stories
by their Human score \(-D\) and encourages the human-authored story to
rank first. Let
\(\mathcal{E}_i=\{\operatorname{Enc}(p_i),\mathbf{0}\}\) denote the
prompt-conditioned and prompt-masked representations for group \(i\).
Every matched group is evaluated under both representations, and the
two modes are optimized jointly:
\[
\mathcal{L}_{\mathrm{energy}}
=
\frac{1}{2N}
\sum_{i=1}^{N}
\sum_{\tilde{e}\in\mathcal{E}_i}
\left[
\mathcal{L}_{\mathrm{point},i}(\tilde{e})
+
\lambda_{\mathrm{list}}
\mathcal{L}_{\mathrm{list},i}(\tilde{e})
\right],
\]
where \(N\) is the number of matched training groups and
\(\lambda_{\mathrm{list}}\) controls the contribution of the auxiliary
listwise objective. Encoder architectures, expanded loss definitions,
and optimization settings are provided in Appendix~D.

For any profile transition \(z\rightarrow z'\), we define its
Human-directed gain as
\[
\Delta D(z\!\rightarrow\!z';\tilde{e}_p)
=
D(z,\tilde{e}_p)-D(z',\tilde{e}_p).
\]
A positive value indicates that the new profile better matches the
Human writing pattern under the same condition. This shared quantity
is used both to select candidate feature transitions during search and
to measure the movement realized after full-story rewriting.

\subsection{Human-Targeted Structured Search}
\label{sec:structured-search}

At inference time, CraftAlign fixes Human as the target and initializes
the search from the predicted profile \(z^{(0)}=\hat{z}\). If
\(D(z^{(0)},\tilde{e}_p)<0\), the initial profile already lies on the
Human side of the decision boundary, so CraftAlign avoids an
unnecessary intervention and generates no additional feature-specific
guidance. Otherwise, CraftAlign constructs single-feature candidates
from the current profile \(z\) and scores each candidate
\(\tilde{z}\) using
\(\Delta D(z\!\rightarrow\!\tilde{z};\tilde{e}_p)\). Because each
candidate changes only one top-level feature, the associated gain also
serves as a local counterfactual sensitivity, making the selected
direction traceable to a specific feature transition.

We construct single-feature candidates by adapting type-aware
structured perturbations for discrete and mixed variables
\citep{schroder2024heat}. Binary values are flipped; categorical
values are replaced by other legal categories; ordinal features and
ordered scales move within their ordered domains; metric scales are
perturbed and projected to legal values; and multi-select features add
or remove one legal option. We retain only schema-valid candidates
whose local transitions are text-actionable and whose net changes from
the initial profile can be rendered through unique canonical guidance
templates. Let
\(\mathcal{C}(z^{(t)};z^{(0)})\) denote the resulting candidates at
step \(t\).

Rather than optimizing toward a predefined target profile, CraftAlign
selects a budgeted high-gain path from multiple schema-valid and
text-actionable revision directions. At each step, it greedily selects
the candidate with the largest Human-directed gain:
\[
z^{(t+1)}
=
\underset{
    \tilde{z}\in\mathcal{C}(z^{(t)};z^{(0)})
}{
    \operatorname{arg\,max}
}
\;
\Delta D
\bigl(
z^{(t)}\!\rightarrow\!\tilde{z};
\tilde{e}_p
\bigr).
\]
The update is accepted only if the maximum gain is positive. The
selected updates form a transition path \(\pi\), and search terminates
when the profile crosses into the Human side, no legal positive-gain
candidate remains, or the budget \(K_{\max}=5\) is exhausted. Multiple
updates to the same feature are permitted but are merged into a single
net transition before guidance rendering. Let \(z^\star\) denote the
final profile. If \(D(z^\star,\tilde{e}_p)<0\), the search has crossed
the model's Human/AI decision boundary; otherwise, \(z^\star\) is the
final greedy state reached under the candidate set and budget. The
search need not cross the boundary or recover a globally optimal or
minimum-length path. Candidate construction, renderability constraints,
pseudocode, and termination details appear in Appendices~E.7 and F.4.

\subsection{Guidance Rendering and Story Revision}
\label{sec:guidance-rendering}

Let \(\pi_i\) denote the transition path selected for story \(x_i\).
Repeated updates to the same feature are first merged into their net
transitions, \(M_i=\operatorname{Merge}(\pi_i)\); a feature that
returns to its initial value contributes no instruction. We maintain a
curated guidance dictionary \(\mathcal{G}\) that maps each renderable
net transition \((k,a\!\rightarrow\!b)\) to a canonical
natural-language revision instruction. The resulting instructions,
ordered by a fixed schema, are appended to a shared base rewrite prompt
to form the final guidance \(h_i\). The number of feature-specific
instructions is \(K_i=|M_i|\). When \(M_i=\varnothing\), the editor
receives only the shared base prompt.

The original story, its optional writing prompt, and the resulting
guidance are then passed to a general-purpose LLM editor, yielding
\(x_i'=\operatorname{Editor}(x_i,p_i,h_i)\). This rendering layer
decouples feature-space direction selection from the choice of editor,
allowing the same selected transitions to guide different editing
models. Because these transitions may concern story-level properties
such as information release, causal organization, character
arrangement, and ending treatment, the editor is instructed to rewrite
the complete story while preserving its premise, main characters,
major events, and approximate length. The guidance identifies revision
priorities rather than enforcing isolated feature changes; their
realization in the revised text is evaluated separately. Appendix~E
describes the guidance dictionary, canonical templates, and rendering
procedure.

\section{Experimental Setup}
\label{sec:experimental-setup}

We use StoryScope, organized into 10,272 prompt-centered groups,
each containing one human-authored story and five AI-generated stories
written for the same prompt, and follow its prompt-level training,
development, and test splits
\citep{russell2026storyscope}. Model selection uses development
prompts, and all reported comparisons use held-out test prompts. All
main downstream experiments use out-of-sample profiles produced by the
frozen feature estimator \(g_\phi\); reference profiles appear only in
RQ1 comparisons and explicitly labeled supplementary ablations.
Experiments follow the CraftAlign pipeline: RQ1 tests whether the local
estimator reproduces the reference profiles while retaining the
information needed downstream; RQ2 evaluates Human/AI writing-pattern
classification under prompt-conditioned and prompt-masked inputs; and
RQ3 tests whether feature directions selected in profile space remain
effective after full-story rewriting and improve perceived
human-likeness.

\subsection{Feature Estimation (RQ1)}
\label{sec:rq1-setup}

StoryScope provides Gemini-3-Flash reference profiles as supervision,
while applying the same feature schema reproducibly to new stories
requires a deployable local estimator. We first measure how closely
\(g_\phi\) reproduces these annotations using metrics matched to each
feature type, reporting results separately for human-authored and
AI-generated stories to assess both overall profile fidelity and
source-specific degradation. We use Macro-F1 for categorical and binary features, quadratic weighted kappa for ordinal features, MAE for metric scales, and Jaccard similarity for multi-select features.

Annotation agreement alone does not reveal whether residual errors
remove information needed by the downstream evaluator. We therefore
train identically configured XGBoost probes on reference and predicted
profiles. Binary Human/AI classification is the main
information-retention probe, while six-way classification over the
human source and five AI generators provides a stricter supplementary
test.

\subsection{Writing-Pattern Evaluation (RQ2)}
\label{sec:rq2-setup}

We evaluate the same test groups under two input settings. In the
\emph{prompt-masked} setting, every available prompt is deliberately
replaced by the all-zero condition vector; we compare XGBoost on
predicted profiles, a separately trained feature-only energy
specialist, and CraftAlign in its Joint-Zero mode. In the
\emph{prompt-conditioned} setting, the original prompt is retained; we
compare XGBoost on predicted profiles concatenated with prompt
embeddings, a separately trained prompt-conditioned energy specialist,
and CraftAlign in its Joint-Prompt mode. Joint-Zero and Joint-Prompt are
two inference modes of the same checkpoint, whereas the specialists are
trained only for their respective settings. The XGBoost baselines test
whether direct classifiers on the same inputs are sufficient, while the
specialists isolate any cost of supporting both input modes in a single
checkpoint.

Macro-F1 is the primary Human/AI classification metric. We additionally report Balanced Accuracy, the mean recall over the Human and AI classes, and AUROC, which measures threshold-independent separation between Human and AI scores. Detailed calculation procedures are provided in the Appendix. All of them indicate the model's classification ability to distinguish between AI-generated and human-written stories. Human Top-1
Accuracy measures the proportion of groups in which the human-authored story receives the highest Human score among the six stories written for the same prompt.

\subsection{Revision Guidance (RQ3)}
\label{sec:rq3-setup}

RQ3 tests the central evaluation-to-revision claim using complementary
model-space and human evidence. The cross-editor analysis examines
whether directions selected in profile space survive full-story
rewriting across different editor models. Because this analysis uses
the same energy evaluator that selects the guidance, a human study
provides an independent assessment of perceived human-likeness.

All revision conditions receive the same base rewrite prompt, original
story, writing prompt, and preservation constraints. They differ only
in their feature-specific guidance. \textbf{Humanize Only} receives no
feature-specific instruction. \textbf{Random Guidance} receives, for
each story, the same number of valid transitions as CraftAlign, sampled
from the same renderable transition space without energy-based
selection and expressed through the same guidance dictionary.
\textbf{CraftAlign Guidance} receives the transitions selected by
Human-targeted energy-guided search. These controls distinguish the
value of targeted direction selection from the effects of generic
rewriting and merely providing additional instructions.

\paragraph{Cross-editor model-space analysis.}
To test whether the guidance generalizes beyond a particular editor, we
sample 50 test prompt--AI-story pairs from distinct prompts, with 10
stories from each AI source, and apply all three revision conditions
using Qwen3.7-Plus, DeepSeek-V4-Pro and GPT-5.6-Luna. For each prompt--story
pair, the mean across the three editors is the primary statistic, so
the prompt--story pair rather than each individual editor output
remains the independent analysis unit. 

For each story, we use the CraftAlign-selected transitions as a fixed
target set shared by all three revision conditions. Target realized is
the fraction of these transitions whose target values appear in the
revised profile. Non-target drift is the fraction of features outside
this target set whose predicted values change from the original profile. We report the averaged movement and supporting diagnostics for each revision
condition, using the matched prompt--story setup to compare CraftAlign with
both revision baselines.
\paragraph{Human evaluation.}
Because CraftAlign and Humanize Only receive identical editor inputs
when the search returns no feature transition, the human study focuses
on 20 guidance-eligible groups with non-empty CraftAlign guidance,
balanced across the five AI sources with four groups per source. The
groups are selected based only on guidance availability and source
balance, before inspecting any revision output or evaluation score.
All revisions in the human study are generated by Gemini-3-Flash,
preventing editor identity from being confounded with revision
condition. Each group contains five anonymized and randomly ordered
stories written for the same prompt: the \textbf{Human Reference}, the
\textbf{Original AI} story, and its \textbf{Humanize Only},
\textbf{Random Guidance}, and \textbf{CraftAlign Guidance} revisions.
The Original AI story tests whether revision improves upon the starting
draft, while the Human Reference serves as a calibration anchor.

Each group is independently evaluated by eight reviewers. After reading
the shared prompt, each reviewer selects exactly two stories that most
resemble natural human-authored writing, without ranking the selected
pair. This protocol avoids imposing a complete ranking and allows
multiple plausible revisions to be recognized. We report the Selection
Rate of all five versions and compare CraftAlign with the Original AI
story and the two revision baselines. For the descriptive figure, we report the mean Selection Rate with reviewer-level min--max ranges.

\section{Results and Analysis}
\label{sec:results}

% We organize the results along the CraftAlign pipeline. RQ1 examines
% whether locally predicted features reproduce the reference annotations
% and retain the information needed for downstream evaluation. RQ2 tests
% whether the resulting class-conditional energy signal supports
% Human/AI writing-pattern classification and within-prompt ranking. RQ3
% then evaluates whether this signal, when converted into feature-level
% revision guidance, produces consistent model-space changes after
% full-story rewriting and improves perceived human-likeness.

\subsection{RQ1: Feature Estimation Reliability}
\label{sec:rq1-results}

\begin{figure}[h]
    \centering
    \includegraphics[width=\columnwidth]
    {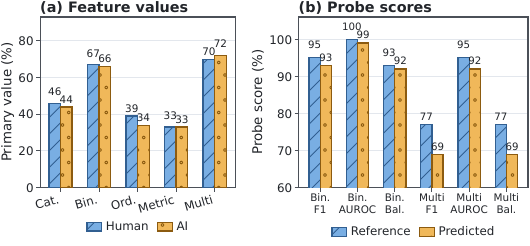}
    \caption{RQ1 feature-estimation reliability. Left: primary scores by
feature type for Human and AI stories. Right: downstream binary
Human/AI and six-way source-classification probe results on reference
and predicted profiles.}
    \label{fig:rq1-diagnostics}
\end{figure}

%  Figure~\ref{fig:rq1-diagnostics} shows that predicted features preserve
% most of the Human/AI discriminative information contained in reference
% annotations. In the binary probe task, reference features achieve the
% best performance with 95\% Macro-F1, while predicted features rank
% second with 93\%, only 2 points lower. AUROC decreases only from 100\%
% to 99\%. For the more challenging six-way classification task, predicted
% features obtain 69\% Macro-F1 and 92\% AUROC, compared with 77\% Macro-F1
% from reference features.

% These results indicate that although predicted features cannot fully
% recover human annotations, they retain the essential information needed
% for downstream Human/AI writing-pattern evaluation. Therefore,
% predicted profiles provide an effective representation for deployment.
Figure~\ref{fig:rq1-diagnostics} shows that the local estimator
provides a stable approximation to the reference feature profiles.
Estimation performance varies across feature types, with binary and
multi-select features achieving the strongest results, but remains
broadly consistent between Human and AI stories. More importantly,
predicted profiles closely approach reference-profile performance on
the primary downstream task: the binary Human/AI probe reaches 93\%
Macro-F1 and 99\% AUROC, compared with 95\% and 100\% using reference
profiles.

Predicted profiles also retain substantial information about finer-grained
source differences, achieving 69\% Macro-F1 and 92\% AUROC on the more
challenging six-way probe. Together, these results show that the local
estimator preserves nearly all information required for the central
Human/AI writing-pattern distinction, while retaining useful signal for
individual source identification. RQ1 therefore validates predicted
profiles as an effective deployment-time representation for the
subsequent energy modeling, feature search, and revision evaluation.

\subsection{RQ2: Human/AI Writing-Pattern Evaluation}
\label{sec:rq2-results}

\begin{table}[!h]
\centering

{\small
\setlength{\tabcolsep}{1mm}
\begin{tabular}{@{}llcccc@{}}
\toprule
Setting & Model & Macro-F1 & AUROC & Bal. Acc. & Top-1 \\
\midrule

\textit{Ref. profile}
& XGBoost
& 94.04 & 99.46 & 94.73 & 99.31 \\

\midrule

\multirow{3}{*}{\textit{No prompt}}
& XGBoost
& 92.59 & \textbf{99.10} & \textbf{94.34} & \textbf{98.47} \\
& Profile-only
& 92.86 & 98.92 & 92.73 & 98.02 \\
& Joint-Zero
& \textbf{93.17} & 98.97 & 93.24 & 98.18 \\

\midrule

\multirow{3}{*}{\textit{With prompt}}
& XGBoost
& 92.39 & 98.99 & 93.77 & \textbf{98.67} \\
& Prompt-cond.
& 93.52 & 99.12 & \textbf{94.09} & 98.31 \\
& Joint-Prompt
& \textbf{93.86} & \textbf{99.19} & 94.08 & 98.18 \\

\bottomrule
\end{tabular}
}

\caption{RQ2 writing-pattern evaluation (\%). Bold marks the
best deployable result within each prompt setting. The
reference-profile result is included for comparison.}
\label{tab:rq2}
\end{table}

\begin{table*}[t]
\centering
\small
\setlength{\tabcolsep}{3.2pt}
\renewcommand{\arraystretch}{1.06}

\begin{tabular*}{0.98\textwidth}{
@{\extracolsep{\fill}}llrrrrrrr@{}
}
\toprule

Editor
& Condition
& \multicolumn{1}{c}{\shortstack[c]{Mean\\$\Delta D_{\mathrm{rev}}\uparrow$}}
& \multicolumn{1}{c}{\shortstack[c]{Positive\\gain $\uparrow$}}
& \multicolumn{1}{c}{\shortstack[c]{H-cross\\$\uparrow$}}
& \multicolumn{1}{c}{\shortstack[c]{Target\\realized $\uparrow$}}
& \multicolumn{1}{c}{\shortstack[c]{Non-target\\drift $\downarrow$}}
& \multicolumn{1}{c}{\shortstack[c]{Length\\valid $\uparrow$}}
& \multicolumn{1}{c}{\shortstack[c]{Win rate\\$\uparrow$}} \\

\midrule

\multirow{3}{*}{GPT-5.6-Luna}
& Humanize only
& -0.55 & 54\% & 22\% & 3\%
& \textbf{12\%} & 86\% & 12\% \\

& Random guidance
& 1.17 & 58\% & 30\% & 47\%
& 14\% & 88\% & 32\% \\

& CraftAlign
& \textbf{2.41} & \textbf{70\%} & \textbf{54\%}
& \textbf{57\%} & 14\% & \textbf{92\%}
& \textbf{56\%} \\

\midrule

\multirow{3}{*}{Qwen3.7-Plus}
& Humanize only
& -1.07 & 40\% & 14\% & 3\%
& \textbf{12\%} & 98\% & 16\% \\

& Random guidance
& -0.74 & 48\% & 24\% & 48\%
& 14\% & \textbf{100\%} & 26\% \\

& CraftAlign
& \textbf{1.64} & \textbf{62\%} & \textbf{38\%}
& \textbf{53\%} & 14\% & \textbf{100\%}
& \textbf{58\%} \\

\midrule

\multirow{3}{*}{DeepSeek-V4-Pro}
& Humanize only
& -1.22 & 40\% & 22\% & 5\%
& \textbf{14\%} & 72\% & 14\% \\

& Random guidance
& 0.28 & 58\% & 30\% & 45\%
& 15\% & \textbf{96\%} & 28\% \\

& CraftAlign
& \textbf{1.95} & \textbf{64\%} & \textbf{50\%}
& \textbf{55\%} & \textbf{14\%} & 88\%
& \textbf{58\%} \\

\midrule

\multirow{3}{*}{\textbf{Avg.}}
& Humanize only
& -0.95 & 45\% & 19\% & 4\%
& \textbf{13\%} & 85\% & 14\% \\

& Random guidance
& 0.24 & 55\% & 28\% & 47\%
& 14\% & \textbf{95\%} & 29\% \\

& CraftAlign
& \textbf{2.00} & \textbf{65\%} & \textbf{47\%}
& \textbf{55\%} & 14\% & 93\%
& \textbf{57\%} \\

\bottomrule
\end{tabular*}

\caption{
RQ3 cross-editor revision results. Target realization and
non-target drift are measured against the CraftAlign-selected target
set for all conditions. Bold marks the best result within each editor;
Avg. reports the macro-average across editors.
% RQ3 cross-editor revision results.
% Positive gain is the percentage of revisions with
% $\Delta D_{\mathrm{rev}}>0$;
% H-cross is the percentage crossing from the AI side to the
% Human side;
% Target realized and Non-target drift measure intended and
% off-target feature changes, respectively;
% Length valid denotes a length ratio within $[0.6,1.4]$;
% and Win rate is the percentage of story--editor cases in which
% a condition obtains the largest $\Delta D_{\mathrm{rev}}$ among
% the three revision conditions.
% Arrows indicate preferred directions, bold marks the best
% result within each editor, and Avg.\ reports the macro-average
% across the three editors.
}
\label{tab:rq3-main}
\end{table*}
% Table~\ref{tab:rq2} compares different approaches under various
% deployment settings. Without prompt information, Joint-Zero achieves
% the best Macro-F1 of 93.17\%, followed by Profile-only with 92.86\%,
% giving Joint-Zero a gain of 0.31 percentage points. When prompt
% information is available, Joint-Prompt obtains the highest Macro-F1 and
% AUROC, reaching 93.86\% and 99.19\%, respectively. It improves over the
% second-best Prompt-conditioned model by 0.34 Macro-F1 points and
% 0.07 AUROC points.

% Moreover, Joint-Prompt is only 0.18 Macro-F1 points behind the
% reference-feature XGBoost upper bound. This demonstrates that CraftAlign
% can achieve near-reference performance using predicted feature profiles,
% confirming that the learned energy function effectively captures
% Human/AI writing-pattern differences.
Table~\ref{tab:rq2} shows that a single jointly trained energy model
supports both deployment settings without an evident performance
trade-off. Joint-Zero and Joint-Prompt achieve the highest Macro-F1 in
their respective settings and slightly outperform the corresponding
single-mode specialists. Prompt conditioning provides a further
0.69-point gain, while Joint-Prompt comes within 0.18 Macro-F1 points
of the reference-profile benchmark. Both joint modes also retain over
98\% Human Top-1 accuracy, indicating that the learned energy function
captures both standalone Human/AI distinctions and within-prompt
relative differences. These results validate it as a stable
prompt-optional scoring function for downstream feature search.

\subsection{RQ3: Revision Guidance Evaluation}
\label{sec:rq3-results}

\begin{figure}[t]
\centering
\scriptsize
\begin{tikzpicture}
\begin{axis}[
    ybar,
    width=\columnwidth,
    height=0.72\columnwidth,
    ymin=0,
    ymax=100,
    ylabel={Selection Rate (\%)},
    symbolic x coords={Human,Original,Humanize,Random,CraftAlign},
    xtick={Human,Original,Humanize,Random,CraftAlign},
    xticklabels={H. Ref.,Original,Humanize,Random,CraftAlign},
    x tick label style={rotate=30,anchor=east},
    bar width=11pt,
    enlarge x limits=0.13,
    ymajorgrids=true,
    grid style={draw=gray!25},
]
\addplot+[draw=black,preaction={fill=green!45},pattern=north east lines,bar shift=0pt] coordinates {(Human,73.8)};
\addplot+[draw=black,preaction={fill=red!45},pattern=horizontal lines,bar shift=0pt] coordinates {(Original,24.4)};
\addplot+[draw=black,preaction={fill=blue!45},pattern=dots,bar shift=0pt] coordinates {(Humanize,30.6)};
\addplot+[draw=black,preaction={fill=orange!55},pattern=crosshatch,bar shift=0pt] coordinates {(Random,25.6)};
\addplot+[draw=black,preaction={fill=purple!55},pattern=north west lines,bar shift=0pt] coordinates {(CraftAlign,45.6)};
\draw[black,line width=0.8pt] (axis cs:Human,30) -- (axis cs:Human,90);
\draw[black,line width=0.8pt] ([xshift=-4pt]axis cs:Human,30) -- ([xshift=4pt]axis cs:Human,30);
\draw[black,line width=0.8pt] ([xshift=-4pt]axis cs:Human,90) -- ([xshift=4pt]axis cs:Human,90);
\draw[black,line width=0.8pt] (axis cs:Original,10) -- (axis cs:Original,40);
\draw[black,line width=0.8pt] ([xshift=-4pt]axis cs:Original,10) -- ([xshift=4pt]axis cs:Original,10);
\draw[black,line width=0.8pt] ([xshift=-4pt]axis cs:Original,40) -- ([xshift=4pt]axis cs:Original,40);
\draw[black,line width=0.8pt] (axis cs:Humanize,15) -- (axis cs:Humanize,65);
\draw[black,line width=0.8pt] ([xshift=-4pt]axis cs:Humanize,15) -- ([xshift=4pt]axis cs:Humanize,15);
\draw[black,line width=0.8pt] ([xshift=-4pt]axis cs:Humanize,65) -- ([xshift=4pt]axis cs:Humanize,65);
\draw[black,line width=0.8pt] (axis cs:Random,5) -- (axis cs:Random,45);
\draw[black,line width=0.8pt] ([xshift=-4pt]axis cs:Random,5) -- ([xshift=4pt]axis cs:Random,5);
\draw[black,line width=0.8pt] ([xshift=-4pt]axis cs:Random,45) -- ([xshift=4pt]axis cs:Random,45);
\draw[black,line width=0.8pt] (axis cs:CraftAlign,35) -- (axis cs:CraftAlign,55);
\draw[black,line width=0.8pt] ([xshift=-4pt]axis cs:CraftAlign,35) -- ([xshift=4pt]axis cs:CraftAlign,35);
\draw[black,line width=0.8pt] ([xshift=-4pt]axis cs:CraftAlign,55) -- ([xshift=4pt]axis cs:CraftAlign,55);
\node[font=\tiny,black,anchor=south] at (axis cs:Human,73.8) {73.8};
\node[font=\tiny,black,anchor=south] at (axis cs:Original,24.4) {24.4};
\node[font=\tiny,black,anchor=south] at (axis cs:Humanize,30.6) {30.6};
\node[font=\tiny,black,anchor=south] at (axis cs:Random,25.6) {25.6};
\node[font=\tiny,black,anchor=south] at (axis cs:CraftAlign,45.6) {45.6};
\end{axis}
\end{tikzpicture}
\caption{
Mean Selection Rates from eight human reviewers. Error bars
show reviewer-level min--max ranges; rates sum to 200\% because each
reviewer selects two stories per group.
% Human evaluation mean Selection Rate. Bars show the mean over eight reviewers, and black error bars show the reviewer-level min-max range. Each reviewer selects two stories per group, so rates across conditions sum to 200\%. Colors and fill patterns jointly distinguish conditions.
}
\label{fig:human-eval-selection}
\end{figure}
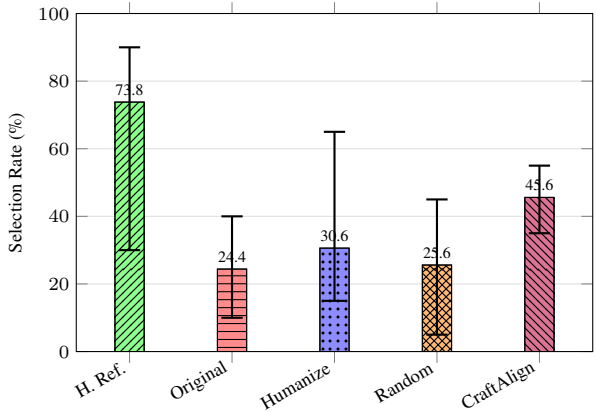

% Table~\ref{tab:rq3-main} and Figure~\ref{fig:human-eval-selection} test the strongest claim of the paper:
% whether feature-level Human/AI contrast can be converted into practical
% revision guidance. 

% \paragraph{Cross-editor model-space analysis.}

% Table~\ref{tab:rq3-main} shows that CraftAlign produces the most consistent
% Human-directed movement across editors. Compared with Humanize Only and
% Random Guidance, its gains indicate that generic polishing or count-matched
% random advice does not reliably move stories toward the target writing
% pattern. By selecting feature transitions through the energy model,
% CraftAlign provides more targeted revision directions while preserving a
% shared editing setup across all conditions.

% \paragraph{Human evaluation.}

% The human evaluation shows that the Human Reference remains the version most often perceived as natural human writing. Although Humanize Only and Random Guidance provide modest improvements, CraftAlign achieves the highest Selection Rate among all non-human versions. Its advantage over Humanize Only and Random Guidance suggests that targeted feature-guided revision is more effective than generic polishing or count-matched random advice. However, CraftAlign still remains below the Human Reference, indicating that it narrows but does not eliminate the perceived gap between AI revisions and human-authored stories.

% Table~\ref{tab:rq3-main} and Figure~\ref{fig:human-eval-selection}
% show that CraftAlign's feature directions remain effective after
% full-story revision and improve perceived human-likeness.

Table~\ref{tab:rq3-main} and
Figure~\ref{fig:human-eval-selection} provide complementary evidence
that the feature-space directions selected by CraftAlign survive
full-story rewriting and translate into improvements perceived by
human readers.

\paragraph{Cross-editor model-space analysis.}
CraftAlign produces positive mean Human-directed movement with all
three editors, whereas Humanize Only moves in the opposite direction
on average for every editor and Random Guidance yields mixed results.
Averaged across editors, CraftAlign achieves positive gain on 65\% of
revisions, crosses to the Human side on 47\%, and obtains the largest
$\Delta D_{\mathrm{rev}}$ in 57\% of story--editor cases. Crucially,
these gains do not come from broader uncontrolled rewriting:
CraftAlign realizes 55\% of the shared target transitions while keeping
non-target drift at 14\%, the same as Random Guidance, and maintaining
a 93\% length-valid rate. The consistent advantage across editors
indicates that the selected directions are not tied to the behavior of
a particular editing model.

\paragraph{Human evaluation.}
Independent human judgments reinforce the model-space results.
CraftAlign reaches a 45.6\% Selection Rate, 15.0 points above the
strongest revision baseline, Humanize Only, and 21.2 points above the
Original AI story. Random Guidance remains close to the original,
showing that merely supplying the same number of valid feature
instructions is insufficient; the benefit comes from selecting
promising transitions through the learned energy signal. More
importantly, the agreement between energy-based movement and human
selection suggests that the improvements are not merely artifacts of
optimizing the evaluator used to select the guidance. The Human
Reference remains highest at 73.8\%. Although the gap to human-authored
stories is not yet fully closed, CraftAlign achieves the strongest
result among all non-human versions and closes a substantial portion
of that gap. Together, these findings highlight the value of explicit
writing features as an actionable bridge between evaluation and
revision, and motivate broader feature-grounded approaches to improving
long-form generation.

\section{Conclusion}

% CraftAlign introduces a feature-grounded framework that extends
% Human/AI story evaluation from source judgment to actionable
% full-story revision. Rather than using explicit writing features
% only for analysis or classification, it treats them as a structured
% intervention space: a shared class-conditional energy signal supports
% both writing-pattern evaluation and the comparison of alternative
% feature transitions, which are then rendered as natural-language
% guidance. Experiments show that locally predicted profiles retain
% most of the discriminative information needed downstream, that the
% learned energy model reliably distinguishes Human and AI writing
% patterns, and that its targeted guidance outperforms generic rewriting
% and count-matched random guidance across multiple editors and in a
% human evaluation. More broadly, explicit craft features can
% provide an auditable interface between evaluation and long-form
% generation. A promising next step is to use feature-grounded
% evaluators as training signals, allowing explicit craft dimensions
% to shape the narrative patterns learned by story-generation models.

CraftAlign moves Human/AI story evaluation beyond source judgment to
actionable full-story revision. It treats explicit writing features as
a structured intervention space, using a shared class-conditional energy
signal both to evaluate writing patterns and to select feature
transitions that are rendered as natural-language guidance. Experiments
show that predicted profiles retain the information required downstream,
the energy model reliably distinguishes Human and AI writing patterns,
and targeted guidance outperforms generic rewriting and count-matched
random guidance across editors and in human evaluation. These
results establish explicit craft features as an auditable interface
between evaluation and long-form generation, and motivate their use as
training signals for future story-generation models.

\clearpage

\bibliography{aaai2027}

% \clearpage
% \include{appendix}

\clearpage
\appendix
\section{A Data, Splits, and Isolation Protocol}

\paragraph{StoryScope data.}
We use the fixed partitions released with StoryScope. The dataset contains
10,272 writing prompts derived from human-authored short stories in Books3.
Each prompt is associated with metadata for the human source story and with
stories generated by up to five language models: GPT-5.4, DeepSeek V3.2,
Kimi K2.5, Gemini 3 Flash, and Claude Sonnet 4.6. Human story text is not
distributed by StoryScope for copyright reasons. Instead, StoryScope releases
source metadata, AI-generated stories, and approximately 61.6K prompt--source
profiles annotated with 304 explicit writing features.

\paragraph{Official partitions.}
We retain the released prompt-level partitions without resampling. The
training, validation, and test sets account for approximately 72.6\%,
13.8\%, and 13.6\% of the retained prompts, respectively. Prompt identifiers
and normalized prompt texts are disjoint across the partitions. The split is
prompt-level rather than author- or anthology-level.

\paragraph{Recovered human-text subset.}
To train the local feature estimator, we join recovered human story texts to
the released Human reference profiles using a one-to-one match on prompt ID.
We retain only examples whose recovered text contains at least 98\% of the
reported original word count. Validation and test retain only confirmed,
directly confirmed, or content-confirmed recoveries, while the expanded
training protocol additionally admits likely and directly-likely recoveries
in the training partition only. In the resulting human-text subset, the
training, validation, and test sets account for approximately 79.5\%,
10.3\%, and 10.2\%, respectively.

\section{B Feature Schema and Evaluation Metrics}

\paragraph{Feature schema.}
We adopt the 304-feature StoryScope schema, which represents writing along ten broad
dimensions: revelation, events, perspective, plot, setting, style, situatedness, temporal structure, agents, and
social networks. The schema contains 124 categorical, 44 binary, 59 ordinal, 45 scale-valued, and 32 multi-select
features. Together, these features describe both surface-level writing style and higher-level narrative properties,
including information release, event and causal organization, point of view, plot structure, temporal arrangement,
characterization, and social relations.

\paragraph{RQ1 feature-estimation metrics.}
RQ1 evaluates whether predicted profiles recover reference StoryScope feature values. Metrics are computed only over
examples with valid reference labels. Each feature is scored separately, and scores are then averaged with equal weight
within each feature type.

For a categorical feature $j$ with $K_j$ legal classes, class-specific F1 is
\[
F1_{j,c}=
\frac{2TP_{j,c}}{2TP_{j,c}+FP_{j,c}+FN_{j,c}},
\]
and the feature-level Macro-F1 is
\[
\operatorname{MacroF1}_j
=
\frac{1}{K_j}\sum_{c=1}^{K_j}F1_{j,c}.
\]
The final categorical score averages over categorical features:
\[
S_{\mathrm{cat}}
=
\frac{1}{|\mathcal F_{\mathrm{cat}}|}
\sum_{j\in\mathcal F_{\mathrm{cat}}}
\operatorname{MacroF1}_j.
\]
All legal classes are included in the average; undefined class F1 values are set to 0.

Binary features use the same Macro-F1 definition with the two classes \texttt{yes} and \texttt{no}:
\[
\operatorname{MacroF1}_j
=
\frac{F1_{j,\mathrm{yes}}+F1_{j,\mathrm{no}}}{2}.
\]
The reported binary score is the equal-weighted average across all binary features.

For ordinal features with ordered labels $0,\ldots,K-1$, we use quadratic-weighted Cohen's $\kappa$. Predicting class
$b$ when the true class is $a$ receives penalty
\[
w_{ab}
=
\frac{(a-b)^2}{(K-1)^2}.
\]
Let $O_{ab}$ be the observed confusion matrix and $E_{ab}$ the expected confusion matrix under independent true and
predicted marginals. The weighted agreement score is
\[
\kappa_w
=
1-
\frac{\sum_{a,b}w_{ab}O_{ab}}
{\sum_{a,b}w_{ab}E_{ab}}.
\]
Each ordinal feature is evaluated separately and then averaged across ordinal features. If $\kappa_w$ is undefined,
for example because labels are constant, it is set to 0 before averaging.

For scale-valued features, we report mean absolute error (MAE) on the taxonomy value scale:
\[
\begin{aligned}
\operatorname{MAE}_j
&=
\frac{1}{N_j}
\sum_{i=1}^{N_j}
\left|\hat v_{ij}-v_{ij}\right|,\\
S_{\mathrm{scale}}
&=
\frac{1}{|\mathcal F_{\mathrm{scale}}|}
\sum_{j\in\mathcal F_{\mathrm{scale}}}
\operatorname{MAE}_j.
\end{aligned}
\]
Scale-valued predictions are mapped back to the taxonomy value scale before MAE
is computed.

For a multi-select feature, let $Y_i$ be the reference option set and $\hat Y_i$ the predicted option set for story
$i$. The sample-wise Jaccard score is
\[
J_i
=
\frac{|Y_i\cap\hat Y_i|}
{|Y_i\cup\hat Y_i|},
\qquad
J_j
=
\frac{1}{N_j}\sum_i J_i.
\]
If both sets are empty, $J_i=1$. The reported multi-select score averages $J_j$ equally over all multi-select
features. Higher values are better for Macro-F1, $\kappa_w$, and Jaccard, while lower MAE is better.

\paragraph{RQ1 source-probe metrics.}
RQ1 also uses diagnostic source probes to evaluate whether predicted profiles retain information about Human/AI and
generator-source identity.
For a class set $\mathcal C$, we compute class-wise precision, recall, and F1 by treating class $c$ as positive and
all other classes as negative:
\[
\begin{aligned}
P_c&=\frac{TP_c}{TP_c+FP_c},\\
R_c&=\frac{TP_c}{TP_c+FN_c},\\
F1_c&=\frac{2TP_c}{2TP_c+FP_c+FN_c}.
\end{aligned}
\]
If the denominator is zero, the corresponding $F1_c$ is set to 0.

For the binary Human/AI probe, the class set is $\mathcal C_{\mathrm{bin}}=\{H,A\}$, where $A$ merges all five AI
sources. The primary metric is
\[
\operatorname{MacroF1}_{\mathrm{bin}}
=
\frac{F1_H+F1_A}{2}.
\]
Balanced accuracy gives equal weight to Human and AI recall:
\[
\operatorname{BalancedAcc}_{\mathrm{bin}}
=
\frac{R_H+R_A}{2}.
\]
Let $s_i=P(y_i=H\mid x_i)$ be the XGBoost Human probability, and let $n_H$ and $n_A$ denote the numbers of Human and
AI examples. Binary AUROC is computed as
\[
\begin{aligned}
\operatorname{AUROC}_{\mathrm{bin}}
&=
\frac{1}{n_Hn_A}
\sum_{i:y_i=H}
\sum_{j:y_j=A}
\Big[
\mathbb{I}(s_i>s_j)\\
&\qquad\qquad
+
\frac{1}{2}\mathbb{I}(s_i=s_j)
\Big].
\end{aligned}
\]
which is equivalent to threshold-independent ROC ranking.

For the six-way source-classification probe,
\[
\mathcal C_6=
\{H,\mathrm{GPT},\mathrm{DeepSeek},\mathrm{Kimi},\mathrm{Gemini},\mathrm{Claude}\}.
\]
Macro-F1 is the equal-weighted average over the six one-vs-rest F1 scores:
\[
\operatorname{MacroF1}_{6}
=
\frac{1}{6}
\sum_{c\in\mathcal C_6}
F1_c.
\]
Six-way AUROC uses macro one-vs-rest averaging. For class $c$, let $s_{ic}=P(y_i=c\mid x_i)$,
$\mathcal P_c=\{i:y_i=c\}$, and $\mathcal N_c=\{j:y_j\neq c\}$. Then
\[
\begin{aligned}
\operatorname{AUC}_c
&=
\frac{1}{|\mathcal P_c||\mathcal N_c|}
\sum_{i\in\mathcal P_c}
\sum_{j\in\mathcal N_c}
\Big[
\mathbb{I}(s_{ic}>s_{jc})\\
&\qquad\qquad
+
\frac{1}{2}\mathbb{I}(s_{ic}=s_{jc})
\Big].
\end{aligned}
\]
and
\[
\operatorname{MacroAUROC}_{6,\mathrm{OvR}}
=
\frac{1}{6}
\sum_{c\in\mathcal C_6}
\operatorname{AUC}_c.
\]
We also compute six-way balanced accuracy as macro recall:
\[
\operatorname{BalancedAcc}_{6}
=
\frac{1}{6}
\sum_{c\in\mathcal C_6}
R_c.
\]

\paragraph{RQ2 writing-pattern evaluation.}
RQ2 evaluates prompt-group Human/AI discrimination. Let Human be the positive class $H$ and AI be the negative class
$A$. Given a Human score $s_i$, the thresholded prediction is
\[
\hat y_i=\mathbb{I}(s_i\geq 0.5).
\]
For class $c\in\{H,A\}$,
\[
F1_c
=
\frac{2TP_c}{2TP_c+FP_c+FN_c}.
\]
The reported Macro-F1 is
\[
\operatorname{MacroF1}
=
\frac{F1_H+F1_A}{2}.
\]
Let $n_H$ and $n_A$ be the numbers of Human and AI examples. AUROC is
\[
\begin{aligned}
\operatorname{AUROC}
&=
\frac{1}{n_Hn_A}
\sum_{i:y_i=H}
\sum_{j:y_j=A}
\Big[
\mathbb{I}(s_i>s_j)\\
&\qquad\qquad
+
\frac{1}{2}\mathbb{I}(s_i=s_j)
\Big],
\end{aligned}
\]
which measures how often a Human story receives a higher Human score than an AI story, with ties counted as 0.5.
Balanced Accuracy, abbreviated as $\operatorname{BalAcc}$ below, is
\[
\begin{aligned}
\operatorname{BalAcc}
&=
\frac{\operatorname{Recall}_H+\operatorname{Recall}_A}{2}
\\
&=
\frac{1}{2}
\left(
\frac{TP_H}{TP_H+FN_H}
+
\frac{TP_A}{TP_A+FN_A}
\right).
\end{aligned}
\]
For Human Top-1 Accuracy, let $h_g$ be the true Human story in prompt group $g$, $m_g=\max_{i\in g}s_i$, and
$T_g=\{i\in g:s_i=m_g\}$ be the set of stories tied for the highest Human score. The group score is
\[
a_g=
\begin{cases}
\dfrac{1}{|T_g|}, & h_g\in T_g,\\[4pt]
0, & h_g\notin T_g,
\end{cases}
\]
and
\[
\operatorname{HumanTop1}
=
\frac{1}{G}
\sum_{g=1}^{G}a_g.
\]
Without ties, this is the fraction of prompt groups in which the Human story ranks first.

\paragraph{RQ3 revision and human-evaluation metrics.}
Let
\[
D(z,e_p)=E_\theta(z,e_p,H)-E_\theta(z,e_p,A),
\]
where lower values indicate greater compatibility with the Human writing pattern. For sample $i$ and revision
condition $m$, let $z_i$ and $z'_{im}$ denote the original and revised profiles. The per-story energy reduction is
\[
\Delta D_{im}
=
D(z_i,e_{p_i})-D(z'_{im},e_{p_i}),
\]
and the condition-level mean is
\[
\overline{\Delta D}_m
=
\frac{1}{N_m}
\sum_{i=1}^{N_m}\Delta D_{im}.
\]
The positive reduction rate is
\[
\operatorname{PRR}_m
=
\frac{1}{N_m}
\sum_{i=1}^{N_m}
\mathbb{I}(\Delta D_{im}>0),
\]
and the Human-side rate is
\[
\operatorname{HSR}_m
=
\frac{1}{N_m}
\sum_{i=1}^{N_m}
\mathbb{I}\!\left[D(z'_{im},e_{p_i})<0\right].
\]

Let $T_{im}$ be the set of target profile coordinates for sample $i$ under condition $m$, and let $v^{*}_{imk}$ be the
target value for coordinate $k$. With tolerance $\epsilon=10^{-5}$, the story-level target realization rate is
\[
r_{im}
=
\frac{1}{|T_{im}|}
\sum_{k\in T_{im}}
\mathbb{I}\!\left(|z'_{imk}-v^{*}_{imk}|\leq\epsilon\right),
\]
and the aggregate target realization rate is
\[
\operatorname{TRR}_m
=
\frac{1}{N_m}
\sum_{i=1}^{N_m}r_{im}.
\]
If $T_{im}$ is empty, $r_{im}$ is set to 0.

For a profile dimension $d$, the story-level non-target drift is
\[
q_{im}
=
\frac{1}{d-|T_{im}|}
\sum_{k\notin T_{im}}
\mathbb{I}\!\left(|z'_{imk}-z_{ik}|>\epsilon\right),
\]
and the aggregate non-target drift is
\[
\operatorname{NTD}_m
=
\frac{1}{N_m}
\sum_{i=1}^{N_m}q_{im}.
\]

Let $W(x)$ be the word count of story $x$. The length ratio and mean length ratio are
\[
\ell_{im}
=
\frac{W(x'_{im})}{W(x_i)},
\qquad
\overline{\ell}_m
=
\frac{1}{N_m}
\sum_{i=1}^{N_m}\ell_{im}.
\]
The length eligibility interval is inclusive:
\[
\operatorname{LER}_m
=
\frac{1}{N_m}
\sum_{i=1}^{N_m}
\mathbb{I}(0.6\leq \ell_{im}\leq 1.4).
\]

For first-place rate, let $\mathcal M=\{\mathrm{Humanize},\mathrm{Random},\mathrm{CraftAlign}\}$. A condition counts
as first only when its $\Delta D$ is the unique largest value for that story:
\[
\operatorname{FPR}_m
=
\frac{1}{N}
\sum_{i=1}^{N}
\mathbb{I}\!\left[
\Delta D_{im}
>
\max_{m'\in\mathcal M\setminus\{m\}}
\Delta D_{im'}
\right].
\]
Ties do not give credit to any condition.

For human evaluation, each reviewer selects two of five anonymized versions per prompt group. The Selection Rate for
version $m$ is
\[
\begin{aligned}
\operatorname{SR}_m
&=
\frac{1}{RG}
\sum_{r=1}^{R}\sum_{g=1}^{G}\\
&\quad
\mathbb{I}\!\left[m\text{ is selected by reviewer }r
\text{ in group }g\right].
\end{aligned}
\]
where $R$ and $G$ denote the numbers of reviewers and evaluation groups. Because each reviewer selects two versions,
$\sum_m\operatorname{SR}_m=2$, i.e., the five Selection Rates sum to $200\%$.

For reviewer-level dispersion, let $\mathcal G_r$ be the groups completed by reviewer $r$ and $G_r=|\mathcal G_r|$.
The reviewer-specific Selection Rate is
\[
\begin{aligned}
\operatorname{SR}_{m,r}
&=
\frac{1}{G_r}
\sum_{g\in\mathcal G_r}\\
&\quad
\mathbb{I}\!\left[m\text{ is selected by reviewer }r
\text{ in group }g\right].
\end{aligned}
\]
The min--max range displayed in the human-evaluation figure is
\[
\operatorname{Range}_m
=
\left[
\min_r\operatorname{SR}_{m,r},
\max_r\operatorname{SR}_{m,r}
\right],
\]
with width $\max_r\operatorname{SR}_{m,r}-\min_r\operatorname{SR}_{m,r}$ when a scalar width is needed.

\section{C Feature Estimator Details}
\label{app:feature-estimator}

\paragraph{Base model and input/output.}
We use the text backbone of Qwen3.5-9B and remove its vision components. The
estimator receives only the story text; the original writing prompt is retained
as metadata but is not provided to the model. Each story is limited to 1,024
tokens using head--tail truncation, retaining 512 tokens from each end.

The hidden state of the last non-padding token is passed through LayerNorm and
dropout before being routed to feature-specific prediction heads. The heads cover
304 feature-specific outputs: 124 categorical, 44 binary, 59 ordinal, 45 scale,
and 32 multi-select features. Categorical features use softmax classification,
binary features use sigmoid classification, and multi-select features use
independent sigmoid outputs over their legal options. Ordinal features and
non-metric ordered scales use a CORN-style ordered objective, whereas metric
scales use Huber regression. Missing labels are masked, and losses are averaged
first within each feature and then across valid features.

\begin{table}[t]
\centering
\small
\begin{tabular}{ll}
\toprule
Setting & Value \\
\midrule
Base model & Qwen3.5-9B text backbone \\
Fine-tuning method & 4-bit QLoRA \\
Quantization & NF4 with double quantization \\
Compute precision & bfloat16 \\
Maximum length & 1,024 tokens \\
Truncation & Head--tail (512/512) \\
LoRA rank & 16 \\
LoRA alpha & 32 \\
LoRA dropout & 0.05 \\
LoRA bias & None \\
Prediction-head dropout & 0.10 \\
Learning rate & $2\times10^{-4}$ \\
Optimizer & AdamW \\
Weight decay & 0.01 \\
Learning-rate schedule & Cosine \\
Warm-up ratio & 0.05 \\
Per-GPU batch size & 1 \\
Gradient accumulation & 16 steps \\
Number of GPUs & 3 \\
Effective global batch size & 48 \\
Training epochs & 3 \\
Optimizer steps & 405 \\
Maximum gradient norm & 1.0 \\
Gradient checkpointing & Enabled \\
Validation interval & 25 steps \\
Early-stopping patience & 4 validations \\
Random seed & 42 \\
Trainable parameters & 47,977,593 \\
\bottomrule
\end{tabular}
\caption{Fine-tuning configuration of the Qwen3.5-9B feature estimator.}
\label{tab:feature-estimator-hyperparameters}
\end{table}

\paragraph{Training, validation, and testing.}
Training stories are shuffled, and the complete validation set is evaluated
every 25 optimizer steps. Checkpoint selection is based exclusively on
validation loss. The selected checkpoint is from step 400, with a validation
loss of 0.4984; it is evaluated once on the held-out test set, obtaining a test
loss of 0.4997.

\section{D Energy Model and Search Procedure Details}
\label{app:energy-search}

\subsection{D.1 Out-of-Sample Profile Generation}
The prompt-level split assigns approximately 79.5\%, 10.3\%, and 10.2\% of the
recovered stories to the training, validation, and test partitions,
respectively, with no prompt overlap between partitions. The selected estimator
is frozen and used to generate predicted profiles for all downstream
experiments.

\paragraph{Class-conditional energy model.}
Let $\hat{z}\in\mathbb{R}^{954}$ denote the encoded predicted profile,
$e_p\in\mathbb{R}^{128}$ the prompt representation, and $y\in\{H,A\}$ a
candidate Human/AI label. The model assigns a scalar energy:
\begin{align}
E_\theta(\hat{z},\tilde{e}_p,y)
&=
f_\theta\!\left(
[\hat{z};\tilde{e}_p;v_y]
\right),\\
D_\theta(\hat{z},\tilde{e}_p)
&=
E_\theta(\hat{z},\tilde{e}_p,H)
-
E_\theta(\hat{z},\tilde{e}_p,A),
\end{align}
where $v_y\in\mathbb{R}^{16}$ is a learned label embedding. Lower energy
indicates greater compatibility with the queried label. Thus, $D<0$ favors
Human and $D>0$ favors AI. The corresponding Human probability is
\[
P_\theta(H\mid\hat{z},\tilde{e}_p)
=
\frac{1}{1+\exp[D_\theta(\hat{z},\tilde{e}_p)]}.
\]

\paragraph{Prompt-conditioned and no-prompt inputs.}
Prompts are encoded using word- and bigram-TF--IDF followed by 128-dimensional
truncated SVD. The two input modes are
\[
\tilde{e}_p=
\begin{cases}
e_p=\operatorname{Enc}(p), & \text{prompt-conditioned},\\
\mathbf{0}_{128}, & \text{no-prompt}.
\end{cases}
\]
The same jointly trained checkpoint is used in both modes; the zero vector
indicates that prompt information is unavailable.

\paragraph{Training objective.}
For a story with source label $y$, the pointwise loss is cross-entropy over the
negative class energies:
\[
\mathcal{L}_{\mathrm{point}}
=
-\log
\frac{\exp[-E_\theta(\hat{z},\tilde{e}_p,y)]}
{\sum_{c\in\{H,A\}}
\exp[-E_\theta(\hat{z},\tilde{e}_p,c)]}.
\]
For each prompt group containing one human story $\hat{z}^{H}$ and five AI
stories $\{\hat{z}^{A}_{j}\}_{j=1}^{5}$, we define the Human score
$s(\hat{z},\tilde{e}_p)=-D_\theta(\hat{z},\tilde{e}_p)$ and use
\[
\mathcal{L}_{\mathrm{list}}
=
-\log
\frac{\exp[s(\hat{z}^{H},\tilde{e}_p)]}
{\exp[s(\hat{z}^{H},\tilde{e}_p)]
+\sum_{j=1}^{5}\exp[s(\hat{z}^{A}_{j},\tilde{e}_p)]}.
\]
The joint objective averages both input modes:
\[
\begin{aligned}
\mathcal{L}_{\mathrm{energy}}
&=
\frac{1}{2N}
\sum_{i=1}^{N}
\sum_{\tilde{e}\in\{e_{p_i},\mathbf{0}\}}
\left[
\mathcal{L}_{\mathrm{point},i}(\tilde{e})
+
\lambda_{\mathrm{list}}
\mathcal{L}_{\mathrm{list},i}(\tilde{e})
\right],\\
\lambda_{\mathrm{list}}
&=0.5.
\end{aligned}
\]

\paragraph{Human-targeted search.}
Starting from $\hat{z}^{(0)}$, each step enumerates schema-valid single-feature
changes in the renderable candidate set. The gain of a candidate $\tilde{z}$ is
\[
G(\tilde{z};\hat{z})
=
D_\theta(\hat{z},e_p)
-
D_\theta(\tilde{z},e_p).
\]
The candidate with the largest positive gain is accepted. Search terminates
when $D<0$, no positive-gain candidate remains, or the five-step budget is
reached. Multiple updates to the same feature are permitted during search and
are merged into a single net transition before guidance rendering.

\begin{table}[t]
\centering
\small
\begin{tabular}{ll}
\toprule
Setting & Value \\
\midrule
Profile dimension & 954 \\
Prompt dimension & 128 \\
Prompt encoder & TF--IDF + truncated SVD \\
TF--IDF features & 20,000 \\
TF--IDF $n$-grams & Word unigrams/bigrams \\
Minimum document frequency & 2 \\
Label-embedding dimension & 16 \\
MLP hidden dimensions & 256, 128 \\
Activation & SiLU \\
Normalization & LayerNorm \\
Dropout & 0.15 \\
Output dimension & 1 scalar energy \\
Optimizer & AdamW \\
Learning rate & $8\times10^{-4}$ \\
Weight decay & $2\times10^{-4}$ \\
Training epochs & 20 \\
Batch size & 256 prompt groups \\
Group composition & 1 Human + 5 AI \\
Listwise-loss weight & 0.5 \\
Checkpoint criterion & Validation Macro-F1 \\
Evaluation seeds & 42, 43, 44 \\
Search model & Joint-Prompt \\
Maximum search steps & 5 \\
Candidate granularity & One feature per step \\
Candidate selection & Maximum positive gain \\
Feature reuse & Allowed; merged \\
\bottomrule
\end{tabular}
\caption{Hyperparameters of the class-conditional energy model and structured search.}
\label{tab:energy-search-hyperparameters}
\end{table}

\section{E Guidance Rendering Templates}
\label{app:guidance-rendering}

\subsection{From Feature Transitions to Natural-Language Guidance}

A selected feature transition is represented as
\[
\tau=(f,v_{\mathrm{current}}\rightarrow v_{\mathrm{target}}),
\]
where $f$ is the feature identifier. Each transition is converted into
natural-language guidance through
\[
\begin{aligned}
\tau
&\longrightarrow
\text{dictionary or taxonomy description}\\
&\longrightarrow
\text{canonical instruction}\\
&\longrightarrow
\text{story-specific edit action}.
\end{aligned}
\]

\subsection{E.7 Renderable Candidate Constraints}
\paragraph{Guidance dictionary.}
The dictionary contains 91 exact transitions covering 72 feature identifiers,
together with five feature-level fallback entries. An exact entry is indexed by
\begin{quote}\small
\texttt{feature\_id ||| current\_value ||| target\_value}
\end{quote}
and contains three fields: \texttt{instruction}, \texttt{preserve}, and
\texttt{avoid}. One example exact transition from the dictionary is:
\begin{tcolorbox}[promptbox,title={Example Exact Transition}]
\textbf{Key:}
\texttt{TMP\_DUR\_011 ||| past ||| mixed\_or\_shifted}

\textbf{Instruction:}
Let the narration occasionally slip out of plain retrospective past tense.
Use a controlled mix of remembered past, immediate present impressions, or
brief reflective commentary when the character is processing an event.

\textbf{Preserve:}
the chronology and all major events.

\textbf{Avoid:}
random tense errors; the shifts should feel motivated by memory, pressure, or
reflection.
\end{tcolorbox}

The canonical verbalization is
\begin{quote}\small
\textbf{Feature:} \texttt{\{feature\_name\}}
(\texttt{\{feature\_id\}}).\\
\textbf{Current profile:} \texttt{\{current\_value\}}.\\
\textbf{Target profile:} \texttt{\{target\_value\}}.\\
\textbf{Revision direction:} \texttt{\{instruction\}}.\\
\textbf{Preserve:} \texttt{\{preserve\}}.\\
\textbf{Avoid:} \texttt{\{avoid\}}.
\end{quote}

If an exact transition is unavailable, the system first uses a feature-level
fallback. Otherwise, it verbalizes the feature definition, the meanings of the
current and target values, observable target-story signals, and common
confusions from the taxonomy. Multi-select transitions are expressed as
``option = selected'' or ``option = not selected''; other feature types use
their schema-defined value labels.

For the final editor input, Gemini-3.5-Flash converts each canonical direction
into conservative, story-specific edit actions using temperature $0$ and a
maximum of 3,500 output tokens. The planner does not rewrite the story. It
specifies \texttt{where\_to\_edit}, \texttt{specific\_action},
\texttt{micro\_examples}, \texttt{must\_preserve},
\texttt{verification\_cues}, and \texttt{avoid}. The selected feature and
target value remain fixed throughout this conversion.

\subsection{Shared Rewrite Prompt}

All three experimental conditions use the following base prompt:
\begin{tcolorbox}[promptbox]
Rewrite the story so it reads more like a polished human-authored short story.

The revised story must contain 85--115\% of the original word count. If
necessary, preserve length by expanding existing scenes rather than summarizing
them.

Do not change the protagonist, named characters, central premise, main plot,
theme, setting logic, scene order, or ending. Do not add a new subplot, main
character, conflict, or moral. Use only local sentence-level revision, dialogue
beats, interiority, scene transitions, sensory detail, pacing, or emphasis.

Do not output analysis, bullet points, metadata, or commentary. Return only the
rewritten story.
\end{tcolorbox}

The original writing prompt and story are appended after this shared
instruction.

\subsection{Condition-Specific Prompts}

\paragraph{Humanize Only.}
No feature-specific guidance is provided:
\begin{tcolorbox}[promptbox]
\texttt{\{shared rewrite prompt\}}\\[2pt]
\textbf{Original prompt:}\\
\texttt{\{original prompt\}}\\[2pt]
\textbf{Original story:}\\
\texttt{\{original story\}}
\end{tcolorbox}

\paragraph{Random Guidance.}
Let $K_i$ be the number of transitions selected by CraftAlign for story $i$.
The baseline samples $K_i$ transitions from the same renderable transition
space without energy-based selection. These transitions use the same
verbalization and story-specific planning procedure as CraftAlign:
\begin{tcolorbox}[promptbox]
\texttt{\{shared rewrite prompt\}}\\[2pt]
\textbf{Use these concrete edit actions while obeying all preservation
rules:}\\
\texttt{\{count-matched random guidance\}}\\[2pt]
\textbf{Original prompt:}\\
\texttt{\{original prompt\}}\\[2pt]
\textbf{Original story:}\\
\texttt{\{original story\}}
\end{tcolorbox}

\paragraph{CraftAlign Guidance.}
The editor receives the transitions selected by the positive-gain energy
search:
\begin{tcolorbox}[promptbox]
\texttt{\{shared rewrite prompt\}}\\[2pt]
\textbf{Use these concrete edit actions while obeying all preservation
rules:}\\
\texttt{\{energy-selected CraftAlign guidance\}}\\[2pt]
\textbf{Original prompt:}\\
\texttt{\{original prompt\}}\\[2pt]
\textbf{Original story:}\\
\texttt{\{original story\}}
\end{tcolorbox}

Random Guidance and CraftAlign Guidance therefore differ only in how feature
transitions are selected. The number of instructions, verbalization procedure,
story-specific planner, and base rewrite prompt are held fixed.

\section{F Revision Evaluation and Human Study Protocol}
\label{app:revision-human-study}

\subsection{Cross-Editor Revision Evaluation}

The cross-editor revision study uses 50 held-out prompt--AI-story pairs from
distinct prompts, with 10 stories from each AI source. For each story, the
CraftAlign-selected transitions define the fixed target set shared by all three
revision conditions. Target realization and non-target drift are computed with
respect to this shared target set, using the matched prompt--story setup to
compare CraftAlign with both revision baselines. The same pool is intended to
support human evaluation, where each group requires reviewers to read five
versions of the same story; using manageable-length stories keeps the reading
burden practical while preserving enough narrative content for meaningful
comparison.

\subsection{F.4 Search Pseudocode and Termination}
Starting from $\hat{z}^{(0)}$, each step enumerates schema-valid single-feature
changes in the renderable candidate set and accepts the candidate with the
largest positive gain. Search terminates when $D<0$, no positive-gain candidate
remains, or the five-step budget is reached. Multiple updates to the same
feature are permitted during search and are merged into a single net transition
before guidance rendering.

\subsection{Human Evaluation Protocol}

% Representative human-evaluation case generated from the blinded reading material and reviewer score sheets.
\paragraph{Representative Case.}
Story 09 is selected as a representative case because CraftAlign Guidance is selected by five of eight reviewers, the highest rate among all non-human versions. The Human Reference serves as a calibration anchor in the full human evaluation, but its text is omitted from this reproduced case for distribution reasons.

\begin{tcolorbox}[promptbox,title={Prompt}]
Write a short story about the trapeze artist whose life is meticulously ordered around his constant residence high in the theater dome, maintaining his art at the pitch of perfection. Explore the moment when his singular focus is irrevocably shattered by an unforeseen, almost irrational anxiety, demanding a second trapeze. Depict the manager's deep unease as he observes the first furrows of care engraving themselves upon the artist's smooth forehead, knowing this new need might only be the beginning of an unmanageable sorrow that threatens his existence. Your story must be approximately 1000 words long.
\end{tcolorbox}

\begin{tcolorbox}[promptbox,title={Original AI}]
The dust in the Th\'e\^atre C\'eleste did not rise to the dome; it settled below, a golden sediment for the footlights, while above, the air remained surgically clear. Aurelian had lived in that vaulted hemisphere for eleven years, inhabiting a network of girders and hemp that only pigeons and riggers knew, ascending to his residence one November afternoon and refusing, with the quiet adamance of a zealot, to descend. His world was a geometry of verticals: the drop of eighty feet to the stage, the climb of thirty feet to the oculus, and the arc of his own flight between, which he maintained at a pitch of perfection so absolute that spectators wept without understanding why. They saw a man become a pendulum, then a comet, then a god severed from the law of gravity. Aurelian saw only mathematics--tensile strength, angular momentum, the exact calculus of release and catch.

His existence was a ritual of excruciating precision. He woke at dawn when the first light pierced the stained glass, filtering through in shafts of garnet and cobalt that he tracked like a sundial. He ate sparingly--food arrived in wicker baskets hauled up by silent stagehands--and he slept on a canvas cot lashed to a catwalk, his body arranged with the same symmetry he brought to the bar. His forehead, when at rest, was a smooth vault of bone and skin, unmarked by horizontal thought, as pristine as the porcelain masks that watched from the theater's proscenium boxes. He had eliminated doubt as rigorously as he had eliminated the ground.

Then came the Tuesday when the air changed.

It happened during a matinee. Aurelian was executing his signature triple pirouette, his body spinning in a tight helix above the orchestra pit, when he reached--not for the bar, which was there, solid and faithful--but for something beyond it. His fingers closed on emptiness. For a fraction of a second, a tremor passed through his frame, invisible to the audience but catastrophic to him. He completed the rotation and caught the trapeze, finishing the sequence to thunderous applause, but as he hung there, chest heaving, he felt a hollowness open inside him, vast and irrational. He looked to his left, into the empty space where the spotlight carved a cone of nothing, and he understood with the clarity of a fever that he needed another bar. A second trapeze. Parallel to his own. Empty. Waiting.

He did not descend to make the request. He wrote it on a slip of rice paper, weighted it with a brass washer, and dropped it onto the stage manager's desk during the intermission lull. It fluttered down like a leaf from a divine tree.

Henri, the manager, read the note and felt his blood slow. He climbed the service ladder--a thing he had not done in years--his heart hammering a dirge against his ribs. He found Aurelian seated on his catwalk, legs dangling into the void, staring at the vacant air with an intensity that bordered on erotic. But Henri's eyes were drawn to the artist's forehead. There, etched with the sudden violence of a crack in marble, ran a single vertical furrow, deep and dark, a line of care that had no business on that serene face.

You work alone, Henri said carefully. His voice did not echo; the dome swallowed sound whole.

Aurelian did not look at him. I require a second apparatus. Identical tensile strength. Identical height. Three meters to the left.

``For a partner?''

No.

``For safety? A net, perhaps?''

No.

The silence between them was heavy with the creak of cooling rigging. Henri understood then, with the gut-certainty of a man who has watched too many artists dissolve into their own obsessions, that this was not a practical demand. It was a symptom. The second trapeze was a confession of insufficiency, an architectural stutter. It represented the possibility of failure, the shadow-self, the double that waits in the wings. Aurelian had spent eleven years proving that one bar was sufficient, that one man could be complete. Now, the smooth forehead was breached, and the furrow there spoke of a fracture in the crystal logic of his isolation.

Henri descended and ordered the installation. He watched from the stalls as the riggers hoisted the second bar, swinging it into place with the reverence of priests installing a relic. It hung there, a mute witness, a dark line against the painted clouds of the ceiling. Aurelian did not touch it that first day, nor the next. He simply performed as before, but now his eyes flickered left, always left, to the empty trapeze that swayed slightly in the ventilation currents. The furrow on his forehead deepened, becoming a trench, a ravine.

Weeks passed. The second trapeze remained pristine, untouched by chalk or sweat. Aurelian began to request other things: a second costume, identical in every stitch, hung beside the first; a second mirror; a second cot. The smooth plane of his face became a topography of worry. Henri watched from below, his own hands trembling as he lit his cigarettes, knowing that this multiplication was only the beginning. The second bar was the first symptom of an unmanageable sorrow, a contagion of doubt that would demand a third trapeze, a fourth, until the dome was crowded with swinging shadows, until the perfection that required solitude collapsed under the weight of its own replication.

One evening, after the audience had gone, Henri climbed halfway up the ladder and stopped. Above him, Aurelian stood poised between the two bars, one hand on his own, one reaching toward the other, frozen in an attitude of eternal indecision. The furrow on his forehead was no longer a line but a canyon, and in it, Henri saw the geography of a fall that had not yet happened, but was now, inevitably, written in the stars.

He descended without speaking, leaving the artist to his high, haunted geometry, knowing that the ground was rising to meet them both, one silent inch at a time.
\end{tcolorbox}

\begin{tcolorbox}[promptbox,title={Humanize Only}]
The dust in the Th\'e\^atre C\'eleste did not rise to the dome; it settled below, a golden sediment for the footlights, while above, the air remained surgically clear. Aurelian had lived in that vaulted hemisphere for eleven years, inhabiting a network of girders and hemp that only pigeons and riggers knew, ascending to his residence one November afternoon and refusing, with the quiet adamance of a zealot, to descend. His world was a geometry of verticals: the drop of eighty feet to the stage, the climb of thirty feet to the oculus, and the arc of his own flight between, which he maintained at a pitch of perfection so absolute that spectators wept without understanding why. They saw a man become a pendulum, then a comet, then a god severed from the law of gravity. Aurelian saw only mathematics--tensile strength, angular momentum, the exact calculus of release and catch.

His existence was a ritual of excruciating precision. He woke at dawn when the first light pierced the stained glass, filtering through in shafts of garnet and cobalt that he tracked like a sundial. He ate sparingly--food arrived in wicker baskets hauled up by silent stagehands--and he slept on a canvas cot lashed to a catwalk, his body arranged with the same symmetry he brought to the bar. His forehead, when at rest, was a smooth vault of bone and skin, unmarked by horizontal thought, as pristine as the porcelain masks that watched from the theater's proscenium boxes. He had eliminated doubt as rigorously as he had eliminated the ground.

Then came the Tuesday when the air changed.

It happened during a matinee. Aurelian was executing his signature triple pirouette, his body spinning in a tight helix above the orchestra pit, when he reached--not for the bar, which was there, solid and faithful--but for something beyond it. His fingers closed on emptiness. For a fraction of a second, a tremor passed through his frame, invisible to the audience but catastrophic to him. He completed the rotation and caught the trapeze, finishing the sequence to thunderous applause, but as he hung there, chest heaving, he felt a hollowness open inside him, vast and irrational. He looked to his left, into the empty space where the spotlight carved a cone of nothing, and he understood with the clarity of a fever that he needed another bar. A second trapeze. Parallel to his own. Empty. Waiting.

He did not climb down to ask in person. Instead, he wrote the request on a slip of rice paper, weighed it with a brass washer, and let it fall onto the stage manager's desk during the quiet of intermission. The note drifted down like a leaf shaken loose from some impossible tree.

Henri, the manager, read the note and felt his blood slow. He climbed the service ladder--a thing he had not done in years--his heart hammering a dirge against his ribs. He found Aurelian seated on his catwalk, legs dangling into the void, staring at the vacant air with an intensity that bordered on erotic. But Henri's eyes were drawn to the artist's forehead. There, etched with the sudden violence of a crack in marble, ran a single vertical furrow, deep and dark, a line of care that had no business on that serene face.

You work alone, Henri said, choosing each word. His voice did not echo; the dome swallowed it before it could reach the rafters.

Aurelian kept his eyes on the empty space. I require a second apparatus. Identical tensile strength. Identical height. Three meters to the left.

``For a partner?''

No.

``For safety? A net, perhaps?''

No, Aurelian said again, and this time the answer sounded almost weary.

The silence between them was heavy with the creak of cooling rigging. Henri understood then, with the gut-certainty of a man who has watched too many artists dissolve into their own obsessions, that this was not a practical demand. It was a symptom. The second trapeze was a confession of insufficiency, an architectural stutter. It represented the possibility of failure, the shadow-self, the double that waits in the wings. Aurelian had spent eleven years proving that one bar was sufficient, that one man could be complete. Now, the smooth forehead was breached, and the furrow there spoke of a fracture in the crystal logic of his isolation.

Henri descended and ordered the installation. He watched from the stalls as the riggers hoisted the second bar, swinging it into place with the reverence of priests installing a relic. It hung there, a mute witness, a dark line against the painted clouds of the ceiling. Aurelian did not touch it that first day, nor the next. He simply performed as before, but now his eyes flickered left, always left, to the empty trapeze that swayed slightly in the ventilation currents. The furrow on his forehead deepened, becoming a trench, a ravine.

Weeks passed. The second trapeze remained pristine, untouched by chalk or sweat, yet Aurelian glanced toward it before every catch. Then he began to request other things: a second costume, identical in every stitch, hung beside the first; a second mirror; a second cot. The smooth plane of his face became a topography of worry. From below, Henri watched his own hands tremble whenever he struck a match, and understood that the multiplication was only beginning. The second bar was the first symptom of an unmanageable sorrow, a contagion of doubt that might demand a third trapeze, a fourth, until the dome filled with swinging shadows and the perfection that required solitude collapsed beneath its own replication.

One evening, after the audience had gone, Henri climbed halfway up the ladder and stopped. Above him, Aurelian stood poised between the two bars, one hand on his own, one reaching toward the other, frozen in an attitude of eternal indecision. The furrow on his forehead was no longer a line but a canyon, and in it, Henri saw the geography of a fall that had not yet happened, but was now, inevitably, written in the stars.

He descended without speaking, leaving the artist to his high, haunted geometry, knowing that the ground was rising to meet them both, one silent inch at a time.
\end{tcolorbox}

\begin{tcolorbox}[promptbox,title={Random Guidance}]
Dust settled below. It gathered beneath the footlights. The dome stayed clear. Golden sediment covered the stage. Above, the air remained surgically clean. Pigeons crossed the girders. Riggers worked there. Aurelian had lived in that vaulted hemisphere for eleven years, inhabiting a network of girders and hemp that only pigeons and riggers knew, ascending to his residence one November afternoon and refusing, with the quiet adamance of a zealot, to descend. His world was a geometry of verticals: the drop of eighty feet to the stage, the climb of thirty feet to the oculus, and the arc of his own flight between, which he maintained at a pitch of perfection so absolute that spectators wept without understanding why. They saw a man become a pendulum, then a comet, then a god severed from the law of gravity. Aurelian saw only mathematics--tensile strength, angular momentum, the exact calculus of release and catch.

His existence followed a precise ritual. He woke at dawn. Light pierced the stained glass. Garnet and cobalt shafts crossed his cot. He tracked them like a sundial. He ate sparingly. Jean the stagehand sent his basket. Luc the rigger checked each rope. Marcelle the costumer brought fresh silks. Dr. Girard checked his pulse weekly. Madame Dupont watched from her box. Henri recorded every request below. Aurelian slept on a canvas cot. It was lashed to the catwalk. His body stayed symmetrical. His forehead remained smooth. It resembled a porcelain mask. He had removed doubt. He had removed the ground.

Then came the Tuesday when the air changed.

It happened during a matinee. Aurelian was executing his signature triple pirouette, his body spinning in a tight helix above the orchestra pit, when he reached--not for the bar, which was there, solid and faithful--but for something beyond it. His fingers closed on emptiness. For a fraction of a second, a tremor passed through his frame, invisible to the audience but catastrophic to him. He completed the rotation and caught the trapeze, finishing the sequence to thunderous applause, but as he hung there, chest heaving, he felt a hollowness open inside him, vast and irrational. He looked to his left, into the empty space where the spotlight carved a cone of nothing, and he understood with the clarity of a fever that he needed another bar. A second trapeze. Parallel to his own. Empty. Waiting.

He did not descend to make the request. He wrote it on a slip of rice paper, weighted it with a brass washer, and dropped it onto the stage manager's desk during the intermission lull. It fluttered down like a leaf from a divine tree.

Henri, the manager, read the note and felt his blood slow. He climbed the service ladder--a thing he had not done in years--his heart hammering a dirge against his ribs. He found Aurelian seated on his catwalk, legs dangling into the void, staring at the vacant air with an intensity that bordered on erotic. But Henri's eyes were drawn to the artist's forehead. There, etched with the sudden violence of a crack in marble, ran a single vertical furrow, deep and dark, a line of care that had no business on that serene face.

You work alone, Henri said carefully. His voice did not echo; the dome swallowed sound whole.

Aurelian did not look at him. I require a second apparatus. Identical tensile strength. Identical height. Three meters to the left.

``For a partner?''

No.

``For safety? A net, perhaps?''

No.

The silence between them was heavy with the creak of cooling rigging. Henri understood then, with the gut-certainty of a man who has watched too many artists dissolve into their own obsessions, that this was not a practical demand. It was a symptom. The second trapeze was a confession of insufficiency, an architectural stutter. It represented the possibility of failure, the shadow-self, the double that waits in the wings. Aurelian had spent eleven years proving that one bar was sufficient, that one man could be complete. Now, the smooth forehead was breached, and the furrow there spoke of a fracture in the crystal logic of his isolation.

Henri descended and ordered the installation. He watched from the stalls as the riggers hoisted the second bar, swinging it into place with the reverence of priests installing a relic. It hung there, a mute witness, a dark line against the painted clouds of the ceiling. Aurelian did not touch it that first day, nor the next. He simply performed as before, but now his eyes flickered left, always left, to the empty trapeze that swayed slightly in the ventilation currents. The furrow on his forehead deepened, becoming a trench, a ravine.

Weeks passed. The second trapeze remained pristine, untouched by chalk or sweat. Aurelian began to request other things: a second costume, identical in every stitch, hung beside the first; a second mirror; a second cot. The smooth plane of his face became a topography of worry. Henri watched from below, his own hands trembling as he lit his cigarettes, knowing that this multiplication was only the beginning. The second bar was the first symptom of an unmanageable sorrow, a contagion of doubt that would demand a third trapeze, a fourth, until the dome was crowded with swinging shadows, until the perfection that required solitude collapsed under the weight of its own replication.

One evening, after the audience had gone, Henri climbed halfway up the ladder and stopped. Above him, Aurelian stood poised between the two bars, one hand on his own, one reaching toward the other, frozen in an attitude of eternal indecision. The furrow on his forehead was no longer a line but a canyon, and in it, Henri saw the geography of a fall that had not yet happened, but was now, inevitably, written in the stars.

He descended without speaking. Aurelian stayed above. The two bars remained beside him. Their lines stayed still. Henri returned below. The routine continued. Aurelian kept his high geometry. Henri watched from the floor. Nothing resolved. The ground remained distant. Each evening repeated the last.
\end{tcolorbox}

\begin{tcolorbox}[promptbox,title={CraftAlign Guidance}]
The dust in the Th\'e\^atre C\'eleste does not rise to the dome; it settles below, a golden sediment for the footlights, while above, the air remains surgically clear, as pristine as the unprinted margins surrounding these words. Aurelian lives in that vaulted hemisphere for eleven years, inhabiting a network of girders and hemp that only pigeons and riggers know, ascending to his residence one November afternoon and refusing, with the quiet adamance of a zealot, to descend. His world is a geometry of verticals: the drop of eighty feet to the stage, the climb of thirty feet to the oculus, and the arc of his own flight between, which he maintains at a pitch of perfection so absolute that spectators weep without understanding why. They see a man become a pendulum, then a comet, then a god severed from the law of gravity. Aurelian sees only mathematics--tensile strength, angular momentum, the exact calculus of release and catch.

His existence is a ritual of excruciating precision. He wakes at dawn when the first light pierces the stained glass, filtering through in shafts of garnet and cobalt that he tracks like a sundial. He eats sparingly--food arrives in wicker baskets hauled up by silent stagehands--and he sleeps on a canvas cot lashed to a catwalk, his body arranged with the same symmetry he brings to the bar. His forehead, when at rest, is a smooth vault of bone and skin, unmarked by horizontal thought, as pristine as the porcelain masks that watch from the theater's proscenium boxes. He has eliminated doubt as rigorously as he has eliminated the ground.

Then came the Tuesday when the air changed.

During a matinee, while executing his signature triple pirouette, his body spinning in a tight helix above the orchestra pit, Aurelian reached--not for the bar, which was there, solid and faithful--but for something beyond it. His fingers closed on emptiness, and, for a fraction of a second, a tremor passed through his frame, invisible to the audience but catastrophic to him. Though he completed the rotation and caught the trapeze, finishing the sequence to thunderous applause, he felt, as he hung there with his chest heaving, a hollowness open inside him, vast and irrational. Into the empty space where the spotlight carved a cone of nothing he looked, and with the clarity of a fever he understood that he needed another bar: a second trapeze, parallel to his own, empty and waiting.

He did not descend to make the request. He wrote it on a slip of rice paper, weighted it with a brass washer, and dropped it onto the stage manager's desk during the intermission lull. It fluttered down like a leaf from a divine tree, landing near the bottom margin of the page.

Henri, the manager, read the note and felt his blood slow. He climbed the service ladder--a thing he had not done in years--his heart hammering a dirge against his ribs. He found Aurelian seated on his catwalk, legs dangling into the void, staring at the vacant air with an intensity that bordered on erotic. But Henri's eyes were drawn to the artist's forehead. There, etched with the sudden violence of a crack in marble, ran a single vertical furrow, deep and dark, a line of care that had no business on that serene face.

You work alone, Henri said carefully. His voice did not echo; the dome swallowed sound whole.

Aurelian did not look at him. I require a second apparatus. Identical tensile strength. Identical height. Three meters to the left.

``For a partner?''

No.

``For safety? A net, perhaps?''

No.

The silence between them was heavy with the creak of cooling rigging. Henri understood then, with the gut-certainty of a man who has watched too many artists dissolve into their own obsessions, that this was not a practical demand. It was a symptom. The second trapeze was a confession of insufficiency, an architectural stutter. It represented the possibility of failure, the shadow-self, the double that waits in the wings. Aurelian had spent eleven years proving that one bar was sufficient, that one man could be complete. Now, the smooth forehead was breached, and the furrow there spoke of a fracture in the crystal logic of his isolation.

Henri descended and ordered the installation. He watched from the stalls as the riggers hoisted the second bar, swinging it into place with the reverence of priests installing a relic. It hung there, a mute witness, a dark line against the painted clouds of the ceiling. Aurelian did not touch it that first day, nor the next. He simply performed as before, but now his eyes flickered left, always left, to the empty trapeze that swayed slightly in the ventilation currents. The furrow on his forehead deepened, becoming a trench, a ravine.

Weeks passed. The second trapeze remained pristine, untouched by chalk or sweat. Aurelian began to request other things: a second costume, identical in every stitch, hung beside the first; a second mirror; a second cot. The smooth plane of his face became a topography of worry. Henri watched from below, his own hands trembling as he lit his cigarettes, knowing that this multiplication was only the beginning. The second bar was the first symptom of an unmanageable sorrow, a contagion of doubt that would demand a third trapeze, a fourth, until the dome was crowded with swinging shadows, until the perfection that required solitude collapsed under the weight of its own replication.

One evening, after the audience had gone, Henri climbed halfway up the ladder and stopped. Above him, Aurelian stood poised between the two bars, one hand on his own, one reaching toward the other, frozen in an attitude of eternal indecision. The furrow on his forehead was no longer a line but a canyon, and in it, Henri saw the geography of a fall that had not yet happened, but was now, inevitably, written in the stars.

He descended without speaking, leaving the artist to his high, haunted geometry, knowing that the ground was rising to meet them both, one silent inch at a time, beyond the edge of the paragraph.
\end{tcolorbox}

\begin{center}
\small
\setlength{\tabcolsep}{3.2pt}
\begin{tabular}{lccccc}
\toprule
Reviewer & H. Ref. & Original & Humanize & Random & CraftAlign \\
\midrule
R1 & 18 & 4 & 4 & 7 & 7 \\
R2 & 14 & 5 & 6 & 5 & 10 \\
R3 & 16 & 7 & 3 & 5 & 9 \\
R4 & 15 & 4 & 6 & 4 & 11 \\
R5 & 14 & 2 & 13 & 1 & 10 \\
R6 & 6 & 8 & 7 & 9 & 10 \\
R7 & 17 & 3 & 6 & 7 & 7 \\
R8 & 18 & 6 & 4 & 3 & 9 \\
\midrule
Mean & 14.8 & 4.9 & 6.1 & 5.1 & 9.1 \\
\bottomrule
\end{tabular}
\captionof{table}{Reviewer-level selection counts over all 20 human-evaluation groups. Each reviewer makes 40 selections in total, selecting two of five versions per group. Mean reports the average count across the eight reviewers; the Human Reference text is not reproduced in the case display.}
\label{tab:human-eval-case-story09}
\end{center}

\end{document}